\RequirePackage{fix-cm}
\documentclass{svjour3}                     % onecolumn (standard format)
\usepackage[margin=1.2in]{geometry}	%margine changes

\smartqed  % flush right qed marks, e.g. at end of proof
\usepackage{graphicx}
\usepackage{multicol}
\usepackage{subcaption}
\usepackage{soul}
\usepackage[table]{xcolor}
\usepackage{hyperref}
\usepackage{multirow}
\usepackage{tabularx}
\usepackage{arydshln}
\usepackage{amssymb}
\usepackage{hyperref}
\usepackage[sort&compress,numbers]{natbib}
\usepackage{natbib}
\usepackage{subcaption}
\usepackage{pdflscape}
\usepackage{longtable}
\usepackage{rotating}
\usepackage{lscape}
\usepackage{adjustbox}
\usepackage{pgf-pie}
\usepackage{lineno}

\journalname{myjournal}

\begin{document}

\title{Text-based automatic personality prediction: A bibliographic review%\thanks{Grants or other notes
}
% \subtitle{%The state-of-the-art in text-based automatic personality prediction\\
%Automatic personality detection: A literature review on texts \\
% A literature review on text-based automatic personality prediction (textual - from text) \\
% %Automatic personality detection; A bibliography review from 2010 \\
% A comparative study on text-based automatic personality prediction \\
% %A systematic review on text-based automatic personality prediction \\
% %Survey on automatic personality detection from texts \\
% %Methods in text-based automatic personality prediction \\
% %Automatic personality detection from NLP point of view \\
% %A bibliography review on automatic personality detection from 2010 \\
% %A comparative study on text-based automatic personality prediction \\
% %A systematic review on text-based automatic personality prediction
% State-of-the-art in text-based automatic personality prediction \\
% Comparative literature review on text-based automatic personality prediction    \\
% A comprehensive review (study) of text-based automatic personality prediction
% }


\author{
        Ali-Reza Feizi-Derakhshi $^1$    \and
        Mohammad-Reza Feizi-Derakhshi $^{*,1}$ \and
        Majid Ramezani$^1$  \and
        Narjes Nikzad-Khasmakhi$^1$ \and
        Meysam Asgari-Chenaghlu$^1$ \and
        Taymaz Akan (Rahkar-Farshi)$^{1,2,3}$    \and
        Mehrdad Ranjbar-Khadivi$^{1,4}$ \and
        Elnaz Zafarni-Moattar$^{1,5}$   \and
        Zoleikha Jahanbakhsh-Naghadeh$^{1,6}$   %\and
        %Mohammad-Ali Balafar$^6$
}

\institute{
            * Corresponding author: Mohammad-Reza Feizi-Derakhshi (\email{mfeizi@tabrizu.ac.ir}) \\
            %\and
            (1) \at
            Computerized Intelligence Systems Laboratory, Department of Computer Engineering, University of Tabriz, Tabriz, Iran \\
            %\and
            (2) \at
            Department of Software Engineering, Ayvansaray University, Istanbul, Turkey.  \\
            %\and
            (3) \at
            Clinical Informatics, Louisiana State University Health Sciences Center Shreveport, Shreveport, USA \\
            %\and
            (4) \at
            Department of Computer Engineering, Shabestar Branch, Islamic Azad University, Shabestar, Iran.   \\
            %\and
            (5) \at
            Department of Computer Engineering, Tabriz Branch, Islamic Azad University, Tabriz, Iran.   \\
            (6) \at
            Department of Computer Engineering, Naghadeh Branch, Islamic Azad University, Naghadeh, Iran.   \\
            %\and
            % \and
            % (6) \at
            % Department of Computer Engineering, University of Tabriz, Iran. \\
            \and
            Ali-Reza Feizi-Derakhshi (\email{derakhshi96@ms.tabrizu.ac.ir, a.f.derakhshi@gmail.com}; ORCID: 0000-0003-3036-1651);\\
            Mohammad-Reza Feizi-Derakhshi (\email{mfeizi@tabrizu.ac.ir}; ORCID: 0000-0002-8548-976X);\\
            Majid Ramezani(\email{m\_ramezani@tabrizu.ac.ir}; ORCID: 0000-0003-0886-7023);\\
    	Narjes Nikzad-Khasmakhi(\email{n.nikzad@tabrizu.ac.ir}; ORCID: 0000-0003-3536-1343);\\
    	Meysam Asgari-Chenaghlu(\email{m.asgari@tabrizu.ac.ir}; ORCID: 0000-0002-7892-9675);\\
    	Taymaz Akan (Rahkar-Farshi)(\email{taymaz.farshi@gmail.com}; ORCID: 0000-0003-4070-1058);\\
            Mehrdad Ranjbar-Khadivi(\email{mehrdad.khadivi@iaushab.ac.ir}; ORCID: 0000-0002-3441-8224);\\
            Elnaz Zafarani-Moattar(\email{e.zafarani@iaut.ac.ir}; ORCID: 0000-0002-2841-4729);\\
            Zoleikha Jahanbakhsh-Nagadeh(\email{zoleikha.jahanbakhsh@iau.ac.ir}; ORCID: 0000-0002-9269-7858);\\
    		%Mohammad-Ali Balafar(\email{balafarila@tabrizu.ac.ir}; ORCID: 0000-0001-5898-0871);
}

\date{Received: date / Accepted: date}

\maketitle

\begin{abstract}
Personality detection is an old topic in psychology and Automatic Personality Prediction (or Perception) (APP) is the automated (computationally) forecasting of the personality on different types of human generated/exchanged contents (such as text, speech, image, video). The principal objective of this study is to offer a shallow (overall) review of natural language processing approaches on APP since 2010. With the advent of deep learning and following it transfer-learning and pre-trained model in NLP, APP research area has been a hot topic, so in this review, methods are categorized into three; pre-trained independent, pre-trained model based, multimodal approaches. Also, to achieve a comprehensive comparison, reported results are informed by datasets.

\end{abstract}

%\linenumbers

%%%%%%%%%%%%%%%%%%%%%%%%%%%%%%%%%%%%----Section----%%%%%%%%%%%%%%%%%%%%%%%%%%%%%%%%%%%%%%%%%%%%%%%%%
\section{Introduction} \label{sec:introduction}
"\textit{The development of language is part of the development of the personality, for words are the natural means of expressing thoughts and establishing understanding between people.}", Maria Montessori.\newline

%The above quotation becomes the basis of what is present in this article, studying natural language processing in individual personality. Personality is defined as the characteristic set of behaviours, cognitions, and emotional patterns \cite{corr2009cambridge} as well as thinking patterns \cite{kazdin2000encyclopedia}, and its external appearance can be seen in writing, speech, decision and other aspects of the social and personal lives of people. Language and written are among the most prominent and accessible signs of personality that represent sights of personality in the absence of people and have always been psychologists' focus. Besides, internet use takes a considerable place in the daily life of everybody, so writings become more available with high speed and verities. Therefore, the involvement of computers in determining the personality of people seems necessary and turned into a study field in computer science.
The above quotation becomes the basis of what is present in this article, studying natural language processing in individual personality. Personality is defined as the characteristic set of behaviours, cognitions, and emotional patterns \cite{corr2009cambridge} as well as thinking patterns \cite{kazdin2000encyclopedia}, and its external appearance can be seen in writing, speech, decision and other aspects of the social and personal lives of people. Language is the most prominent and the most available aspects of individuals' personality. Meanwhile, written text is one of the most utilized appearance of language. Developing the Internet based infrastructure, such as social media, e-mails, and different texting contexts, have made the language appearance of people more available. Consequently, considering the increase in internet based communications, it would be so exciting to become aware of individuals' personality, inspite of their absence. Therefore, the involvement of computers in determining the personality of people seems necessary and turned into a study field in computer science.

\textit{Personality} refers to a person's long-term set of characteristics and behaviours \cite{Bergner2020}. It encompasses people's moods, attitudes, and ideas, and it manifests itself in their relationships with others \cite{peters2015psychology}. In general, it encompasses all of the behavioural features (both inherent and acquired) that may be noticed in people's social interactions and even their interactions with their surroundings. The word personality comes from the Latin \textit{persona}, which refers to a mask worn by an actor in ancient Greek and Roman plays to symbolize and convey the character's personality attributes.(please refer to \cite{Schultz2016,Ewen2010,Eysenck1998} for more details)

Automatic Personality Prediction (or Perception) (APP) is the automatic prediction of the personality of individuals and usually done by computers. With the increasing variety of data types available for analysing the personality of people, aspects of view to APP increases likewise. In this point of view to the assortment of APP, data types can be named as: speech \cite{Jothilakshmi2017,Su2018,Gilpin2018,Mohammadi2012}, image \cite{Sang2016,Allen2016,Chaudhari2019,Lokhande2017}, video \cite{Kindiroglu2017,Aslan2019}, text \cite{Ramezani2020,Han2020,Xue2021}, social media activities \cite{Zhu2020,tadesse_2018,Lima2014}, touch screen interaction \cite{kuster2018,roy_roy_sinha_2018}, and so on. Also, each of these has subsets and divisions of text-based APP which can be mentioned are email \cite{Shen2013}, SMS \cite{Yakoub2015}, and tweets \& posts on social media \cite{Arnoux2017}. Thereby, the key standpoint of this study is analysing APP methods through text data type and NLP.

Personality should be measured and classified to make it more comparative, and this goes back to psychology. Psychologist put forward many personality trait models; such as Allport's trait theory \cite{Allport1937}, Cattell's 16 Factor Model \cite{Cattell1970} (Table \ref{tbl:cattel16-overview} shows characteristics of traits), Eysenck Personality Questionnaire(EPQ) \cite{Eysenck1975}, Myers-Briggs Type Indicator (MBTI) \cite{Briggs1976}, and Big Five \cite{P.John1999}. Among these, two models; MBTI and Big Five are popular and widely used models, especially in APPs. MBTI has four main dimensions, Introversion versus Extraversion (I-E), Sensing versus iNtuiting (S-N), Thinking versus Feeling (T-F), and Judging versus Perceiving (J-P), that each people categorized in two dimensions. Figure \ref{fig:mbti-overview} defines each MBTI dimension characteristics. The second popular personality model is Big Five. This model consists of five traits, and people may get in one or more trait. Also, two different approaches are taken thus binary modelling (0 and 1 for each trait) or continuous modelling (each trait get a value in range 0 to 1); those two approaches are being used in APP datasets. Openness, Conscientiousness, Extroversion, Agreeableness, and Neuroticism are traits of Big Five which are called OCEAN in abbreviative. Table \ref{tbl:big_five_overview} illustrates characteristics in each OCEAN trait. Without any doubt, from the psychological perspective, there are several drawbacks in current APP methods that basically refer to the existing datasets like, assigning binary labels to individuals’ personality traits which are considered as a continuum. Obviously, having comprehensive datasets would remove such drawbacks and improve the performance of APP.

\begin{table}
	\scriptsize
	{
		\begin{center}
			\caption{An overview of Big Five personality traits model \cite{Ramezani2020}.}
			\label{tbl:big_five_overview}
			\fontsize{7}{8}
			\begin{adjustbox}{max width=\textwidth}
				\begin{tabular}{@{}p{5cm}@{}p{3.2cm}@{}p{5cm}@{}} % <-- Alignments: 1st column left, 2nd middle and 3rd right, with vertical lines in between
					
					\hline
					\hline
					\cellcolor{gray!15}	\textbf{Description of \textit{LOW} values \textcolor{gray} {-}} 	\textcolor{gray} {$\infty$ $\longleftarrow$}  	&	 \cellcolor{gray!15}	\textbf{Personality trait} \centering	&  		\cellcolor{gray!15}	\textcolor{gray} {$\longrightarrow$ +$\infty$}	 \textbf{ Description of \textit{HIGH} values}\\
					\hline
					\hline
					%--------   Openness ---------
					& \textbf{Openness (O)} \centering & 	
					\\
					\hline
					
					\tiny
					\begin{tabular}{p{5cm}}
						
							•	Dislikes changes \\		
							•	Does not enjoy new things \\
							•	Conventional\\
							•	Resists new ideas\\	
							•	Prefers familiarity\\
							•	Not very imaginative\\
							•	Has trouble with abstract or theoretical concepts\\
							•	Skeptical\\
							•	Traditional in thinking\\
							•	Consistent and cautious
					\end{tabular}
					
					&&
					\tiny
					\begin{tabular}{p{5cm}}
							•	Very creative\\	
							•	Clever, insightful, daring, and varied interests\\
							•	Embraces trying new things or visiting new places\\
							•	Unconventional\\
							•	Focused on tackling new challenges\\
							•	Intellectually curious\\
							•	Inventive\\
							•	Happy to think about abstract concepts\\
							•	Enjoys the art\\
							•	Eager to meet new people
					\end{tabular}
					
					\\
					\hline
					
					%--------   Conscientiousness ---------
					& 	\textbf{Conscientiousness (C)} 	\centering	 & 
					
					\\
					\hline
					
					\tiny
					\begin{tabular}{p{5cm}}
							•	Easy going and careless\\
							•	Messy and less detailed-oriented\\
							•	Dislikes structure and schedule\\
							•	Fails to return things or put them back, where they belong\\
							•	Procrastinates on important tasks and rarely completes them on time\\
							•	Fails to stick to a schedule\\
							•	Is always late when meeting others
					\end{tabular}
					
					& & 
					
					\begin{tabular}{p{5cm}}
						\tiny		%  To minimize the font size
						\begin{minipage}[t]{\columnwidth}		% In order to Wrap text and evoide line exceedings to next cells
							•	Competent and efficient\\
							•	Goal- and detail-oriented\\
							•	Well organized, self-discipline and dutiful\\
							•	Spends time preparing\\
							•	Predictable and deliberate\\
							•	Finishes important tasks on time\\
							•	Does not give in to impulses\\
							•	Enjoys adhering to a schedule\\
							•	Is on time when meeting others\\
							•	Works hardly\\ 
							•	Reliable and resourceful\\
							•	Persevered
						\end{minipage}				% In order to Wrap text and evoide line exceedings to next cells
					\end{tabular}
					
					\\
					
					\hline
					
					%--------   Extraversion (E) ---------
					&	 \textbf{Extroversion (E)} 		\centering	 & 
					\\
					\hline
					
					\tiny
					\begin{tabular}{p{5cm}}
							•	Introspective\\
							•	Solitary and reserved\\
							•	Dislikes being at the center of attentions\\
							•	Feels exhausted when having to socialize a lot\\
							•	Finds it difficult to start conversations\\
							•	Dislikes making small talks\\
							•	Carefully thinks things before speaking\\
							•	Thoughtful
					\end{tabular}
					
					& & 
					
					\tiny
					\begin{tabular}{p{5cm}}
							•	Outgoing and energetic\\
							•	Assertive and talkative\\
							•	Able to be articulate\\
							•	Enjoys being the center of attentions\\
							•	Likes to start conversations\\
							•	Enjoys being with others and meeting new people\\
							•	Tendency to be affectionate\\
							•	Finds it easy to make new friends\\
							•	Has a wide social circle of friends and acquaintances\\
							•	Says things before thinking about them\\
							•	Feels organized when around other people\\
							•	Social confidence
					\end{tabular}
					
					\\
					
					\hline
					
					%--------   Agreeableness (A) ---------
					&	 \textbf{Agreeableness (A)}  \centering		& 
					\\
					\hline
					
					\tiny
					\begin{tabular}{p{5cm}}
							•	Challenging and detached\\
							•	Takes little interest in others\\
							•	Can be seen as insulting or dismissive of others\\
							•	Does not care about other people's feelings or problems\\
							•	Can be manipulative\\
							•	Prefers to be competitive and stubborn\\
							•	Insults and belittles others\\
					\end{tabular}
					
					& & 
					
					\tiny
					\begin{tabular}{p{5cm}}
							•	Friendly and compassionate toward others\\ 
							•	Altruist and unselfish\\
							•	Loyal and patient\\
							•	Has a great deal of interest in and wants to help others\\
							•	Feels empathy and concern for other people\\
							•	Prefers to cooperate and be helpful\\
							•	Polite and trustworthy\\
							•	Cheerful and considerate\\
							•	Modest
					\end{tabular}
					
					\\
					
					\hline
					
					%--------   Neuroticism (N) ---------
					& 		\textbf{Neuroticism (N)} 		\centering		 & 
					\\
					\hline
					
					\tiny
					\begin{tabular}{p{5cm}}
							•	Emotionally stable\\
							•	Deals well with stress\\
							•	Rarely feels sad or depressed\\
							•	Does not worry much and is very relax\\
							•	Confident and secure\\
							•	Optimist
					\end{tabular}
					
					& & 
					\tiny
					\begin{tabular}{p{5cm}}
						
							•	Anxious of many different things and nervous\\
							•	Experiences a lot of stress\\
							•	Irritable\\
							•	Impulsive and moody\\
							•	Jealous\\
							•	Lack of confidence\\
							•	Self-criticism\\
							•	Oversensitive\\
							•	Instable and insecure\\
							•	Timid\\
							•	Pessimist
						\end{tabular}
					
					\\
					\hline
				\end{tabular}
			\end{adjustbox}
		\end{center}
	}
\end{table}

Automatic personality prediction is a young research subject in computer science and artificial intelligence and, in recent years, has become a very active research subject. Further research is demanded to maximize the impact of automated personality prediction, as this is still in its early stages. This review will seek to provide an overview and comprehensive view of the different types of text-based automatic personality prediction methods. The motivation for this review is based on the following:
\begin{itemize}
    \item Personality classification is still in the early stages of development and requires further investigation. As a challenging and complex issue, investigating additional directions for future research is essential to enrich extant personality classification techniques further.
    \item The need for a bibliographic literature review is observed after investigating the gradual research regarding personality classification. Hence, the present work is based on the bibliographic literature review.
    \item This review has been motivated by the rapid advances in personality classification, so the researchers have identified, summarized and evaluated the relevant studies in this field.
\end{itemize}

In recent years, the NLP field is faced with a revolution that APP does not get its benefits. In this study, APP articles from 2010, which involve textual inputs, reviewed in three categories: classical text representation and feature extraction methods, articles assisted novel pre-trained word representations, and methods with multimodal approaches (besides text, other data types included).

The rest of this study is organized in the following manner: in section \ref{sec:NLP-materials}, materials and methods which are used as the baselines of APP studies introduced briefly. Section \ref{sec:methods} are the overview of methods and consist of three sub-sections. The results of studies are structured based on datasets and are shown in section \ref{sec:evaluation}; the datasets are also explained in this section. Finally, some concluding remarks are given in section \ref{sec:conclusion}.
%%%%%%%%%%%%%%%%%%%%%%%%%%%%%%%%%%%%----Section----%%%%%%%%%%%%%%%%%%%%%%%%%%%%%%%%%%%%%%%%%%%%%%%%%
\section{NLP materials in personality prediction} \label{sec:NLP-materials}
The approach of this study is to overview researches in APP, which conducted on texts. To this end, material and methods of text analysis are given in this section briefly. Linguistic Inquiry and Word Count (LIWC) is one the most used and developed tools of APP and in section \ref{subsec:LIWC} has been reported. The next material based on NLP is called MRC and is a dictionary psycholinguistic in English. The last one is about Embedding techniques that represent words for text analysis, typically in the form of a real-valued vector that encodes the meaning of the word such that the words that are closer in the vector space are expected to be similar in meaning.
%%%%%%%%%%%%%%%%%%%%%%%%%%%%%%%%%%----Sub-Section----%%%%%%%%%%%%%%%%%%%%%%%%%%%%%%%%%%%%%%%%%%%%%%%
%\subsection{N-grams}

%%%%%%%%%%%%%%%%%%%%%%%%%%%%%%%%%%----Sub-Section----%%%%%%%%%%%%%%%%%%%%%%%%%%%%%%%%%%%%%%%%%%%%%%%
\subsection{Linguistic Inquiry and Word Count (LIWC)} \label{subsec:LIWC}
LIWC (Linguistic Inquiry and Word Count) is introduced by \citet{Pennebaker1996} and developed in years \cite{Pennebaker1999,Tausczik2010} as NLP tools for the psychological purpose. LIWC is a text analysis tool that provides statistical reports that are very useful in determining texts to aim for emotional and cognitive analysis of people. Since 2001, two updated versions of LIWC are introduced in 2007 and 2015. In each version some features were added, and table \ref{tbl:LIWC-comparison} show all features reported by LIWC and the deferences between the last two versions. LIWC reports consist of 91 features in 15 categories. This test could be done online in \url{https://liwc.wpengine.com/}.

\begin{table}[]
    \centering
    \caption{Comparison of LIWC 2015 and LIWC 2007 \cite{LIWC_Comparison}.}
    \label{tbl:LIWC-comparison}
    \tiny
    \begin{tabular}{m{3cm}m{1.7cm}m{1.8cm}m{1.8cm}m{1.8cm}}
        \hline
        \textbf{LIWC Dimension} &\textbf{Output Label}  &\textbf{LIWC2015 Mean} &\textbf{LIWC2007 Mean}   &\textbf{LIWC 2015/2007 Correlation}    \\
        \hline
        Word Count  &WC &11,921.82  &11,852.99	&1.00   \\
        \rowcolor{cyan!10} \multicolumn{5}{c}{\textit{Summary Variable}}   \\
        Analytical Thinking	&Analytic	&56.34	&	&	\\
        \rowcolor{gray!10} Clout	&Clout	&57.95	&	&	\\
        Authentic	&Authentic	&49.17	&	&	\\
        \rowcolor{gray!10} Emotional Tone	&Tone	&54.22	&   &   \\
        \rowcolor{cyan!10} \multicolumn{5}{c}{\textit{Language Metrics}}   \\
        Words per sentence	&WPS	&17.40	&25.07	&0.74	\\
        \rowcolor{gray!10} Words $>$ 6 letters	&Sixltr	&15.60	&15.89	&0.98	\\
        Dictionary words	&Dic	&85.18	&83.95	&0.94	\\
        \rowcolor{cyan!10} \multicolumn{5}{c}{\textit{Function Words}}   \\
        Function Words	&function	&51.87	&54.29	&0.95	\\
        \rowcolor{gray!10} \hspace{1mm}Total pronouns	&pronoun	&15.22	&14.99	&0.99	\\
        \hspace{2mm}Personal pronouns	&ppron	&9.95	&9.83	&0.99	\\
        \rowcolor{gray!10} \hspace{3mm}1st pers singular	&i	&4.99	&4.97	&1.00	\\
        \hspace{3mm}1st pers plural	&we	&0.72	&0.72	&1.00	\\
        \rowcolor{gray!10} \hspace{3mm}2nd person	&you	&1.70	&1.61	&0.98	\\
        \hspace{3mm}3rd pers singular	&shehe	&1.88	&1.87	&1.00	\\
        \rowcolor{gray!10} \hspace{3mm}3rd pers plural	&they	&0.66	&0.66	&0.99	\\
        \hspace{2mm}Impersonal pronouns	&ipron	&5.26	&5.17	&0.99	\\
        \rowcolor{gray!10} Articles	&article	&6.51	&6.53	&0.99	\\
        Prepositions	&prep	&12.93	&12.59	&0.96	\\
        \rowcolor{gray!10} Auxiliary verbs	&auxverb	&8.53	&8.82	&0.96	\\
        Common adverbs	&adverb	&5.27	&4.83	&0.97	\\
        \rowcolor{gray!10} Conjunctions	&conj	&5.90	&5.87	&0.99	\\
        Negations	&negate	&1.66	&1.72	&0.96	\\
        \rowcolor{cyan!10} \multicolumn{5}{c}{\textit{Grammar Other}}   \\
        Regular verbs	&verb	&16.44	&15.26	&0.72	\\
        \rowcolor{gray!10} Adjectives	&adj	&4.49	&	&	\\
        Comparatives	&compare	&2.23	&	&	\\
        \rowcolor{gray!10} Interrogatives	&interrog	&1.61	&	&	\\
        Numbers	&number	&2.12	&1.98	&0.98	\\
        \rowcolor{gray!10} Quantifiers	&quant	&2.02	&2.48	&0.88	\\
        \rowcolor{cyan!10} \multicolumn{5}{c}{\textit{Affect Words}}   \\
        Affect Words	&affect	&5.57	&5.63	&0.96	\\
        \rowcolor{gray!10} \hspace{1mm} Positive emotion	&posemo	&3.67	&3.75	&0.96	\\
        \hspace{1mm} Negative emotion	&negemo	&1.84	&1.83	&0.96	\\
        \rowcolor{gray!10} \hspace{2mm} Anxiety	&anx	&0.31	&0.33	&0.94	\\
        \hspace{2mm} Anger	&anger	&0.54	&0.60	&0.97	\\
        \rowcolor{gray!10} \hspace{2mm} Sadness	&sad	&0.41	&0.39	&0.92	\\
        \rowcolor{cyan!10} \multicolumn{5}{c}{\textit{Social Words}}   \\
        Social Words	&social	&9.74	&9.36	&0.96	\\
        \rowcolor{gray!10}  \hspace{1mm} Family	&family	&0.44	&0.38	&0.94	\\
        \hspace{1mm} Friends	&friend	&0.36	&0.23	&0.78	\\
        \rowcolor{gray!10} \hspace{1mm} Female referents	&female	&0.98	&	&	\\
        \hspace{1mm} Male referents	&male	&1.65	&	&	\\
        \rowcolor{cyan!10} \multicolumn{5}{c}{\textit{Cognitive Processes}}   \\
        Cognitive Processes &cogproc	&10.61	&14.99	&0.84	\\
        \rowcolor{gray!10} \hspace{1mm} Insight    &insight	&2.16	&2.13	&0.98	\\
        \hspace{1mm} Cause	&cause	&1.40	&1.41	&0.97	\\
        \rowcolor{gray!10} \hspace{1mm} Discrepancies	&discrep	&1.44	&1.45	&0.99	\\
        \hspace{1mm} Tentativeness	&tentat	&2.52	&2.42	&0.98	\\
        \rowcolor{gray!10} \hspace{1mm} Certainty	&certain	&1.35	&1.27	&0.92	\\
        \hspace{1mm} Differentiation    &differ	&2.99	&2.48	&0.85	\\
        \rowcolor{cyan!10} \multicolumn{5}{c}{\textit{Perpetual Processes}}   \\
        Perpetual Processes	&percept	&2.70	&2.36	&0.92	\\
        \rowcolor{gray!10} \hspace{1mm} Seeing	&see	&1.08	&0.87	&0.88	\\
        \hspace{1mm} Hearing	&hear	&0.83	&0.73	&0.94	\\
        \rowcolor{gray!10} \hspace{1mm} Feeling	&feel	&0.64	&0.62	&0.92	\\
        \rowcolor{cyan!10} \multicolumn{5}{c}{\textit{Biological Processes}}   \\
        Biological Processes	&bio	&2.03	&1.88	&0.94	\\
        \rowcolor{gray!10} \hspace{1mm} Body	&body	&0.69	&0.68	&0.96	\\
        \hspace{1mm} Health/illness	&health	&0.59	&0.53	&0.87	\\
        \rowcolor{gray!10} \hspace{1mm} Sexuality	&sexual	&0.13	&0.28	&0.76	\\
        \hspace{1mm} Ingesting	&ingest	&0.57	&0.46	&0.94	\\
        \rowcolor{cyan!10} \multicolumn{5}{c}{\textit{Core Drives and Needs}}   \\
        Core Drives and Needs	&drives	&6.93	&	&	\\
        \rowcolor{gray!10} \hspace{1mm} Affiliation	&affiliation	&2.05	&   &	\\
        \hspace{1mm} Achievement	&achieve	&1.30	&1.56	&0.93	\\
        \rowcolor{gray!10} \hspace{1mm} Power	 &power	&2.35	&	&	\\
        \hspace{1mm} Reward focus	&reward	&1.46	&	&	\\
        \rowcolor{gray!10} \hspace{1mm} Risk/prevention focus	&risk	&0.47	&   &   \\
        \rowcolor{cyan!10} \multicolumn{5}{c}{\textit{Time Orientation}}   \\
        Past focus	&focuspast	&4.64	&4.14	&0.97	\\
        \rowcolor{gray!10} Present focus	&focuspresent	&9.96	&8.10	&0.92	\\
        Future focus	&focusfuture	&1.42	&1.00	&0.63	\\
        \rowcolor{cyan!10} \multicolumn{5}{c}{\textit{Relativity}}   \\
        Relativity	&relativ	&14.26	&13.87	&0.98	\\
        \hspace{1mm} Motion	&motion	&2.15	&2.06	&0.93	\\
        \rowcolor{gray!10} \hspace{1mm} Space	&space	&6.89	&6.17	&0.96	\\
        \hspace{1mm} Time	&time	&5.46	&5.79	&0.94	\\
        \rowcolor{cyan!10} \multicolumn{5}{c}{\textit{Personal Concerns}}   \\
        Work	&work	&2.56	&2.27	&0.97	\\
        \rowcolor{gray!10} Leisure	&leisure	&1.35	&1.37	&0.95	\\
        Home	&home	&0.55	&0.56	&0.99	\\
        \rowcolor{gray!10} Money	&money	&0.68	&0.70	&0.97	\\
        Religion	&relig	&0.28	&0.32	&0.96	\\
        \rowcolor{gray!10} Death	&death	&0.16	&0.16	&0.96	\\
        \rowcolor{cyan!10} \multicolumn{5}{c}{\textit{Informal Speech}}   \\
        Informal Speech	informal	&2.52	&	&	\\
        \rowcolor{gray!10} \hspace{1mm} Swear words	&swear	&0.21	&0.17	&0.89	\\
        \hspace{1mm} Netspeak	&netspeak	&0.97	&	&	\\
        \rowcolor{gray!10} \hspace{1mm} Assent	&assent	&0.95	&1.11	&0.68	\\
        \hspace{1mm} Nonfluencies	&nonfl	&0.54	&0.30	&0.84	\\
        \rowcolor{gray!10} \hspace{1mm} Fillers	&filler	&0.11	&0.40	&0.29	\\
        \rowcolor{cyan!10} \multicolumn{5}{c}{\textit{All Punctuation}}   \\
        Periods	&Period	&7.46	&7.91	&0.98	\\
        \rowcolor{gray!10} \hspace{1mm} Commas	&Comma	&4.73	&4.81	&0.98	\\
        \hspace{1mm} Colons	&Colon	&0.63	&0.63	&1.00	\\
        \rowcolor{gray!10} \hspace{1mm} Semicolons	&SemiC	&0.30	&0.24	&0.98	\\
        \hspace{1mm} Question marks	&QMark	&0.58	&0.95	&1.00	\\
        \rowcolor{gray!10} \hspace{1mm} Exclamation marks	&Exclam	&1.0	&0.91	&1.00	\\
        \hspace{1mm} Dashes	&Dash	&1.19	&1.38	&0.98	\\
        \rowcolor{gray!10} \hspace{1mm} Quotation marks	&Quote	&1.19	&1.38	&0.76	\\
        \hspace{1mm} Apostrophes	&Apostro	&2.13	&2.83	&0.76	\\
        \rowcolor{gray!10} \hspace{1mm} Parentheses (pairs)	&Parenth	&0.52	&0.25	&0.90	\\
        \hspace{1mm} Other punctuation	&OtherP	&0.72	&1.38	&0.98	\\
    \end{tabular}
\end{table}

\citet{Mairesse2007} developed a method based on LIWC on the Essays dataset. The authors fed the Essays dataset to LIWC, and the outputs contain LIWC features with a personality trait label in the Big Five dimension. Since the LIWC queries are not free, Mairesse dataset deploys as LIWC features in most research. Mairesse developed a framework and is available in \url{http://farm2.user.srcf.net/research/personality/recognizer}, but it should be noted that it is a while that the framework did not support.
%%%%%%%%%%%%%%%%%%%%%%%%%%%%%%%%%%----Sub-Section----%%%%%%%%%%%%%%%%%%%%%%%%%%%%%%%%%%%%%%%%%%%%%%%
\subsection{MRC Psycholinguistic Database} \label{subsec:MRC}
MRC is a publicly available machine usable dictionary that includes different (up to 26) linguistic and psycholinguistic attributes for 150,837 English words. Different semantic, syntactic, phonological, and orthographic details about the words have made it suitable for miscellaneous purposes of researches in psychology, linguistics and artificial intelligence. Word association data are also included in the database. The first version was introduced in 1981 \cite{Coltheart1981} and the second and last version \cite{wilson_1988} is now available in \url{https://websites.psychology.uwa.edu.au/school/mrcdatabase/mrc2.html} with the details and statistics.
%%%%%%%%%%%%%%%%%%%%%%%%%%%%%%%%%%----Sub-Section----%%%%%%%%%%%%%%%%%%%%%%%%%%%%%%%%%%%%%%%%%%%%%%%
\subsection{Embedding techniques} \label{subsec:embeddings}
Any input should model to understand by computer and writings are no exception, so embeddings duty is that. The smallest meaningful segment of writing is words, and that is why fundamental is word embedding in this task. Typically, each word or token represent by a vector. So basic embedding is one-hot thus a dictionary of words generate then each word represented by a vector that only one cell's value is one and other are zero, and vector size equals to dictionary size. In this representation, vectors are orthogonal, so there is no semantics and relation between words. Moreover, in large corpuses vector size get large and make need storage to save and handle it.

The problems and disabilities of one-hot vector generate a felt need for new embeddings. Word2Vec \cite{Mikolov2013,Mikolov2013a} was the first word embedding able to map words to vectors considering to semantic. Word2Vec became the cornerstone of other embedding techniques with their facilities. FastText \cite{Bojanowski2017} and GloVe \cite{pennington2014glove} are examples in evolution word embedding techniques. All of the embeddings should be trained, so there are some pre-trained language models (PLMs) with differences in trained database and attitude of training (CBOW and Skip grams).

Transformers \cite{vaswani2017attention} \footnote{More information on \cite{minaee2020deep}} introduced in 2018, made a revolution in embedding techniques thus make more parallelism possible than other architectures (such as CNNs and RNNs). Hence, computers get able to train larger models, since large-scale Transformer-based PLMs appeared. The most well-known transformer-based PLM could be named BERT \cite{Devlin2018}. Based on BBERT, numerous models arise by a different point of view; namely, RoBERTa \cite{Liu2019} (robust and larger), AlBERT \cite{lan2020albert} (high-speed training and lower memory), DistilBERT \cite{sanh2020distilbert} (40\% smaller and 60\% faster).
%%%%%%%%%%%%%%%%%%%%%%%%%%%%%%%%----Sub-Section----%%%%%%%%%%%%%%%%%%%%%%%%%%%%%%%%%%%%%%%%%%%%%
%\subsection{Classifiers} \label{}

%\begin{landscape}
% \begin{table}
%     \caption{Text}
%     \label{tab:text}
%     \centering
%     \begin{tabular}{lllll}
%         \hline
%         Ref & Trait Theory  & Type of Input Data    & Technique & Dataset   \\
%         \hline
%         %\endhead
%         1   & 2 & 3 & 4 & 5 \\
%         \noalign{\smallskip}\hline
%     \end{tabular}
% \end{table}
%\end{landscape}

%%%%%%%%%%%%%%%%%%%%%%%%%%%%%%%%%%%%----Section----%%%%%%%%%%%%%%%%%%%%%%%%%%%%%%%%%%%%%%%%%%%%%%%%%
\section{Methods (Overview)} \label{sec:methods}
In natural language processing, representation of input is the most important component, and the smallest part of a text is words. In recent years novel representations called pre-trained word embeddings have been got trends and make revolutions in text mining. APP is not unaffected, and novel methods based on word embedding techniques have appeared. In reviewing APPs, at first methods without pre-trained word embedding techniques describes, secondly pre-trained word embedding based methods details, and at the end methods with more than text input introduces.

% \begin{figure}
%     \centering
%     \begin{tikzpicture}
%         \pie[rotate=270, text=pin, color={blue!30, red!40, orange!70}]  
%      {63.5/Conferences, 29.4/Journals, 7.1/ArXiV}
%     \end{tikzpicture} 
%     \caption{Caption}
%     \label{fig:my_label}
% \end{figure}

%%%%%%%%%%%%%%%%%%%%%%%%%%%%%%%%%%----Sub-Section----%%%%%%%%%%%%%%%%%%%%%%%%%%%%%%%%%%%%%%%%%%%%%%%
\subsection{PLM free APPs} \label{subsec:no-word-embedding}

\citet{Poria2013} proposed a combinational algorithm for detecting personality using LIWC and MRC on textual features. 81 LIWC features plus 26 MRC features extracted from the Essays dataset. The proposed EmoSenticSpace was a novel representation method on a graph of EmoSenticNet \cite{havasi2007conceptnet} and fed to a blending algorithm. The output is a 100-dimensional vector and by "bag of concept", averaging done to vectors to represent a text. The authors trained five Sequential Minimal Optimization (SMO) classifier, one for each of the big five traits.

\citet{Verhoeven2013}, proposed an ensemble model for recognizing personality from Facebook. The authors trained three SVM classifiers, the first one trained on Facebook. The second is trained on the Essays dataset, and as the meta classifier, the output of two classifiers is fed to the third SVM classifier.

Part of speech (POS) is a basic feature of texts. In \cite{Wright2014}, authors used POS \& POS n-grams of texts beside bag of words, word sentiment, negations, and vocabulary size features to predicting personality. As a classifier, SVM is deployed in two formats; 2-class and 3-class. 2588 university students essays collected on the Big Five personality scale to evaluate the method.

LIWC consists of 19 features, and \citet{Tighe2016} tried to reduce feature size for the purpose of achieving better results on the Essays. In this way, Information Gain and Principal Component Analysis (PCA) feature reduction techniques have been examined. It is proved that some LIWC features do not have adequate information on Essays for APP.

The early word embedding models, e.g. Word2Vec, have some common practical problems that make using them hard and somehow restricted. The first one is unseen words; if a word does not see in the learning stage, make the model in trouble. The second problem occurs once you want to train a model, and there are too many parameters. Due to these problems, \citet{FLiu2016} proposed an embedding model to embed personality texts. C2W2S4PT (Character to Word to Sentence for Personality Traits) is a three-stage Bi-RNN based model; firstly, characters modelling; secondly, word modelling based on the first stage; third, sentence embedding using words representation using feedforward neural network. In figure \ref{fig:FLiu2016} the proposed architecture is illustrated. In \cite{Liu2017} the proposed C2W2S4PT evaluated on English, Spanish and Italian languages to proving language independency of C2W2S4PT.

%\textcolor{red}{Methods depended to a language: Indonesia (\cite{Ong2017}, \cite{Lukito2017})}

In \cite{Zheng2019} research, to take advantage of huge unlabeled data, Pseudo Multi-view Co-training (PMC) \cite{chen2011automatic} is adopted, an effective Semi-supervised learning algorithm, to build a personality prediction model. To extract adequate linguistic features, both LIWC and n-gram, along with the Word2Vec word embedding technique, is trained on mypersonality dataset to predict personality through textual data. Figure \ref{fig:Zheng2019} illustrates the overall framework of method.

Personality2Vec \cite{Guan2020} is the name of a user-personality embedding technique through a user has generated texts. Semantic and linguistic features of texts constructs graph and a biased walk strategy has been proposed to divide users into groups with maximum similar personality users in the same group. As shown in fig \ref{fig:Guan2020}, in the first linguistic and semantic features extract in order to pass into the learning part. Linguistic features are 103-dimension of LIWC and 10-dimension of special linguistic features, which have been proposed by the authors. 

Paper \cite{Sun2020}, deployed network representation learning (NRL) as the novelty and for the first time in APP. AdaWalk generates the graph of documents on personality datasets in two approaches: classification and regression. NRL presented node(words or token) by using SkipGram.

Another graph-based approach called personality graph convolutional networks (personality GCN) introduced by \citet{Wang2020_Encoding}. The authors aimed to create a graph to model users, documents, and words by the core of the co-occurrence of words in a document. The weight of edges is calculated by TF-IDF for a document-word edge and pointwise mutual information (PMI) for a word-word edge. The classification layer is the last one, five classifier for Big Five traits. Figure \ref{fig:Wang2020_Encoding} illustrate an overview of the three layers of GCN. It is worthwhile to say that words, users, and documents represented by a one-hot vector.

\begin{figure}
    \centering
    \includegraphics[width=0.7\textwidth]{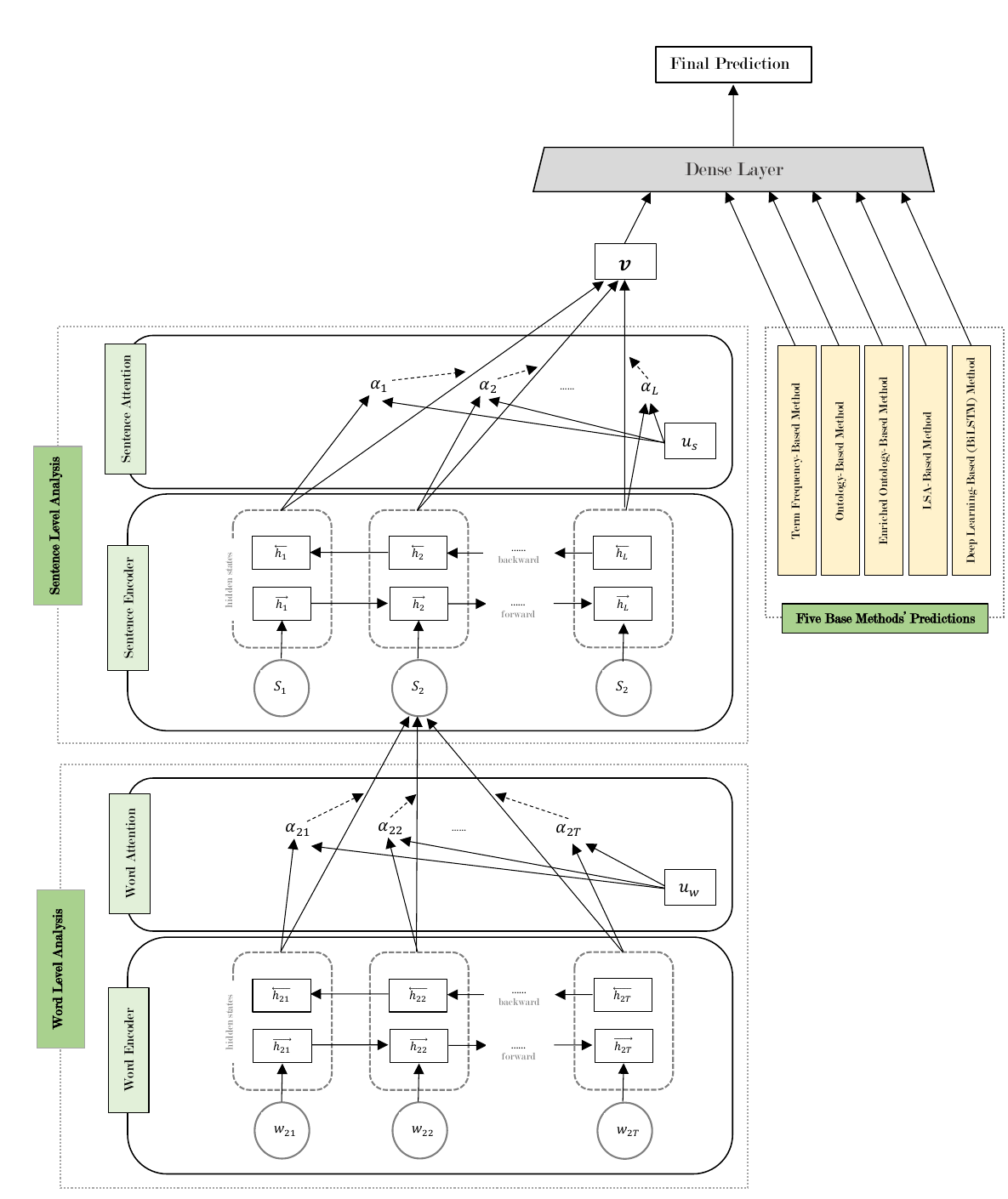}
    \caption{The architecture of a Hierarchical Attention Network (HAN) combined with base
    methods' predictions (ensemble) to carry out stacking in document level classification proposed by \cite{Ramezani2020}.}
    \label{fig:ramezani-overview}
\end{figure}

One of the most recent research in this area proposed five new APP methods: term frequency vector-based, ontology-based, enriched ontology-based, latent semantic analysis (LSA)-based, and deep learning-based (BiLSTM), by a contribution of enhancing accuracy \cite{Ramezani2020}. These five introduced models used as the base to hierarchical attention network (HAN) ensemble model. The authors evaluated the methods on the Essays dataset and achieved enhancing accuracy on the ensemble method. The architecture of the proposed HAN stacking model is shown in figure \ref{fig:ramezani-overview}.

Application of knowledgeable representation of the input text in a graph structure (namely knowledge graph) is a powerful technique in artificial intelligence, and Ramezani et al. \cite{Ramezani_Graph-Enabled} deployed this technique in APP for the first time. The proposed Knowledge graph is a set of interlinked descriptions of concepts that were built by matching the input text’s concepts with DBpedia knowledge base entries. The graph was enriched with the DBpedia ontology, NRC Emotion Intensity Lexicon, and MRC psycholinguistic database to achieve a more robust knowledge representation (Figure \ref{fig:ramezani-2} shows the proposed archtecture and deployment of knowledge graph). The knowledge graph, which served as a knowledgable alternative to the input text, was then embedded to produce an embedding matrix. Finally, the resultant embedding matrix was given to four different deep learning models, including convolutional neural networks (CNN), basic recurrent neural networks (RNN), long short term memory (LSTM), and bidirectional long short term memory (Bidirectional LSTM) (BiLSTM).

Ramezani et al. \cite{Ramezani_KGrAt-Net} in the follow of knowledge graph-based approaches, deployed KGrAt-Net, which is a Knowledge Graph Attention Network text classifier, for the first time to perform Automatic Personality Prediction (APP). As figure \ref{fig:ramezani-3} shows, the proposed approach consists of three stages: pre-processing, knowledge representation, and KGrAt-Net classifier.
%%%%%%%%%%%%%%%%%%%%%%%%%%%%%%%%%%----Sub-Section----%%%%%%%%%%%%%%%%%%%%%%%%%%%%%%%%%%%%%%%%%%%%%%%
\subsection{PLM-based APPs} \label{subsec:plm-based-methods}
The first research that deployed a PLM to APP is done by \citet{Majumder2017} based on Word2Vec. This research deployed two approaches, Document-level using Mairesse features and Word-level using Word2vec and CNN model to achieve 300 dimension representation of words to model sentences and documents, respectively, by n-gram sigh of view. Both approaches' representations concatenated to fed to a fully connected layer for classification (Figure \ref{fig:mjmuder-2017} illustrate the architecture of the proposed method). Five models, one for each of five traits, trained on the Essays dataset.

2CLSTM is the name of the model proposed by \citet{Sun2018} that tries to learn structural features based on latent sentence group (LSG). At the first step, each word embedded into a 100-d dimension vector through GloVe pre-trained model. 2LSTM encodes the vectors into sentences that pass to the next section. The following sections have to model relationships of sentences, and LSG does this. Latent Sentence Group (LSG) is defined as a synthesis that consists of a number of sentence vectors which are closely connected in some coordinates. To do this, CNN networks deployed to learn 1,2,3-grams. Figure \ref{fig:2CLSTM_Sun2018} illustrates the layers and schema of 2CLSTM. A dense layer and max-pooling layer follow immediately to generate the final vector to pass into a classifier which there was softmax.

In \cite{Darliansyah2019}, a personality prediction methods by using sentiment of short texts is introduced by naming SENTIPEDE. The aim was to determine personality in Big Five by using textual features with sentiment features of short texts which the base was Twitter. The sentiment of text compute by Valence Aware Dictionary and Sentiment Reasoner (VADER) \cite{hutto2014vader}, a rule-based framework for sentiment classification in English, to gain a label from positive, negative, or nonpartisan. Then, GloVe word embedding is deployed to vectorise words to pass to a CNN-LSTM model along with the sentiment labels. The case study of the paper was Uber and related tweets.

In introducing a new dataset, some base methods and evaluation of them by a standard dataset are necessary. FriendPersona is the name of a new dataset introduced by \citet{Jiang2020}. In \cite{Jiang2020}, five models have been developed, which names are: ABCNN (CNN with attention mechanism), ABLSTM (Bidirectional LSTM attention mechanism), HAN (Hierarchical Attention Network), BERT, RoBERTa. BERT and RoBERTa as a PLM are fine-tuned on both datasets, Essays and FriendPersona.

In paper \cite{Mehta2020}, two approaches have been studied; personality prediction based on psycholinguistic features and language model features. In both approaches, features extracted from texts are fed to a classifier, SVM and MLP, to classify texts into personality traits. Mairesse, SenticNet \cite{AAAI1816839}, NRC Emotion Lexicon \cite{mohammad2013crowdsourcing}, VAD Lexicon \cite{mohammad-2018-obtaining}, and Readability\footnote{A number of calculated readability mea- sures based on simple surface characteristics of the text. These measures are basically linear regressions based on the number of words, syllables, and sentences.} are the extracted psychological features. On the other hand, BERT PLM is deployed as a language model features extractor.

\citet{Kazameini2020} is segmented documents into 250 tokens length sub-documents in order to feed to BERT(base) PLM. Based on the essays dataset used in this paper, documents length is 650 tokens on average, so four layers of [CLS] concatenated with 84 Mairesse features as features of a document. SVM is the classifier of the proposed method, which is trained on sub-documents in parallel mode like a bagged classifier, and at last, the final trait is predicted by majority voting. Figure \ref{fig:Kazameini2020} shows the authors proposed method.

One of the recent APP methods combined semantic and emotional features in order to determine personality trait from multi-text \cite{Ren2021}. On the semantic side, BERT is deployed to vectorize texts and using a self-attention mechanism, sentence-level representation is generated. On the other side, SenticNet5 \cite{AAAI1816839} has extracted the sentiment of the sentences to map to a vector. Both vectors are concatenated and fed to a classification network. CNN, GRU and LSTM are different neural networks that trained to label personality trait.

SEPRNN (semantic-enhanced personality recognition neural network) \cite{Xue2021} is proposed with the goal of avoiding dependency to feature selection in APP and modelling semantic from word-level representations. GloVe plm is deployed to vectorize words, then a BiGRU model learned to extract left and right context of words, but since semantics did not consider. To capture the higher level of semantic representations from contexted textual data, vectors are fed to a fully connected network to text modelling of documents. In the end, a fully connected network with sigmoid activation function is adopted to learn a two-dimensional vector for binary classification of personality.An illustration of the proposed method is displayed in figure \ref{fig:Xue2021}.

Transformer-MD (multi-document Transformer) is the name of the method proposed by \citet{Yang2021}. The core of the proposed method is to put together information in multiple posts to represent an overall personality for each user. Authors tried to solve two problems: post-orders bias into posts on personality and individually posts processing of a person to personality detection. In order to this, encoding each post by Transformer-MD allows access to information in the other posts of a user through Transformer-XL’s memory tokens that share the position embedding. For the cases of multi-trait personality detection, a dimension attention mechanism on top of Transformer-MD was set. The overview of the proposed methods is shown in figure \ref{fig:Yang2021}.
% \begin{figure}
%     \centering
%     \includegraphics[width=0.8\textwidth]{AttentiveDATN.pdf}
%     \caption{DATN \cite{}}
%     \label{fig:asgari-overview}
% \end{figure}
%%%%%%%%%%%%%%%%%%%%%%%%%%%%%%%%%%----Sub-Section----%%%%%%%%%%%%%%%%%%%%%%%%%%%%%%%%%%%%%%%%%%%%%%%
\subsection{Multimodal APPs} \label{subsec:multimodal-methods}
One of the first deep learning used methods in multimodal approaches deployed text in couple with authors information to achieve the writer's personality \cite{Yu2017}. According to the limitations and obstacles of pre-trained word embedding methods in 2017, word2vec, the authors trained their word embedding model based on skip-gram using the mypersonality dataset. In the trained model, modelling words position was not taken into account, so applying N-grams seemed the best approach. In order to apply this approach, two approaches were taken, CNN-based and Bi-RNN-based. Figure \ref{fig:Yu2017} shows the architecture of \cite{Yu2017}. After the word modelling part, the author's information is 7-D vector concatenate and goes on. Each of the five personality traits has been gotten trained in a neural network.

\citet{Farnadi2013} explored the use of machine learning techniques for inferring a user’s personality traits from their Facebook status updates. four kinds of numeric features:
\begin{itemize}
    \item LIWC features
    \item Social Network features: 7 features related to the social network of the user: (1) network size, (2) betweenness, (3) nbetweenness, (4) density, (5) brokerage, (6) nbroker- age, and (7) transitivity.
    \item Time-related features: 6 features related to the time of the status updates (we assume that all the times are based on one time zone): (1) frequency of status updates per day, (2) number of statuses posted between 6-11 am, (3) num- ber of statuses posted between 11-16, (4) number of sta- tuses posted between 16-21, (5) number of statuses posted between 21-00, and (6) number of statuses posted be- tween 00-6 am.
    \item Other features: 6 features not yet included in the categories above: (1) total number of statuses per user, (2) number of capitalized words, (3) number of capital let- ters, (4) number of words that are used more than once, (5) number of urls, and (6) number of occurrences of the string PROPNAME— a string used in the data to replace proper names of persons for anonymisation purposes.
    \end{itemize}
Different concatenation of features is analysed with 3 classification algorithms named SVM, KNN, and Naive Bayes.

\citet{Wei2017} proposed a comprehensive view to  APP multimodal approach that accompanied texts, avatars, and emojis. Pearson correlation, Text-CNN, and Bag-of-Word (BOW) clusters are the textual-based features extracted from the Weibo tweets collected in the research. Pearson correlation computed between words and the personality traits to selecting top 2000 words are strongly correlated to personality. In order to LIWC limited capability in representing users linguistic patterns in short and informal texts, 1,500 Chinese words and all the punctuations in the bag-of-words format clustered using the k-means algorithm and then the count the number of items within each cluster used instead of LIWC. As the last feature, a convolutional architecture called Text-CNN trained to models words in vector form in reference to \cite{kim-2014-convolutional} model. The structure of the proposed algorithm is shown in figure \ref{fig:Wei2017}. As seen, each type of inputs lasts for a classifier ((Logistic Regression) to specify the trait. As the final step, a stacked ensemble algorithm, generalization-based ensemble method, attached to the classifiers as the final result classifier.

It is common to utilize acoustically features simultaneously with transcripts of speech to achieve the personality of people. In \cite{An2018}, four features demonstrated in order to predict personality, namely, acoustic-prosodic low-level descriptor features (LLD), LIWC, Dictionary of Affect in Language (DAL), and word embeddings. DAL features extracted using Whissell’s Dictionary of Affect in Language (DAL), and 19 features extracted in this research. In the word embedding part, Google’s pre-trained skip-gram vectors and Stanford’s pre-trained GloVe has been used. Two approaches have been adopted for modelling documents based on embedding vectors, averaging and LSTM neural network. One strange thing in the proposed method is that both of PLMs vectors are fed to the model and concatenated with three other features. Moreover, two strategies proposed by the authors, first, concatenate features and then five fully-connected layers end with five neurons as the classifier; second, each of five features fed to a three fully-connected layers block before concatenating and then fed to five neurons for classification similarly.

\section{Evaluating methods} \label{sec:evaluation}
Every proposed method should be evaluated to prove its performance. In APP evaluations consist of two parts, dataset is the first and metric of assessment is the second part. Five datasets are available in text-based APP, and metrics vary on the concept of evaluation. This section consists of two parts; part one details the datasets and part two defines results of methods in datasets and metrics.
%%%%%%%%%%%%%%%%%%%%%%%%%%%%%%%%%%----Sub-Section----%%%%%%%%%%%%%%%%%%%%%%%%%%%%%%%%%%%%%%%%%%%%%%%
\subsection{Datasets} \label{subsec:dataset}
Each method should evaluate and compare it with other methods, and this requires a fair condition. To achieving fair compare, ground truths should be same including metrics and datasets. In this part, five datasets, Essays, MyPersonality, YouTube, FriendPersona, and Kaggle MBTI are benchmark in text-based APPs are introduced.

%%%%%%%%%%%%%%%%%%%%%%%%%%%%%%%%----Sub-sub-Section----%%%%%%%%%%%%%%%%%%%%%%%%%%%%%%%%%%%%%%%%%%%%%
\subsubsection*{Essays}
Essays (also called stream-of-consciousness essays) is the first and most cited text dataset in Automatic personality prediction. The dataset introduced by \citet{Pennebaker1999} that consist of 2468 anonymous essays in English annotated in Big Five scale. The dataset annotated in two modes; classification and regression. Thus each essay has two Big Five values; first, each trait has a binary value; second, each trait is real value in the aim of regression. The Essay mainly deployed in classification purpose and table \ref{tbl:essay_label_dist} shows the number of essays in each trait. In should be notated that one row of data was an error and dismissed in the table's values. Moreover, the distribution of essay's components reported in table \ref{tbl:essay_dataset_elements}.

\begin{table}[]
    \centering
    \caption{Distribution of the Essays dataset labels.}
    \label{tbl:essay_label_dist}
    \begin{tabular}{cccccc}
        \hline
        \textbf{Label}   & \textbf{Opn} & \textbf{Con} & \textbf{Ext} & \textbf{Agr} & \textbf{Neu}    \\
        \hline
        Yes &1271  &1253  &1276  &1310  &1233  \\
        No  &1196  &1214  &1191  &1157  &1234  \\
        \hline
    \end{tabular}
\end{table}

\begin{table}[]
    \centering
    \caption{The elements count of Essay dataset.}
    \label{tbl:essay_dataset_elements}
    \begin{tabular}{c|cccccc}
    \hline
    & {\small \textbf{Sentences}}    & {\small \textbf{Characters}}   & {\small \textbf{Spaces}}   & {\small \textbf{Words}}   & {\small \textbf{Punctuations}}   & {\small \textbf{Stop words}}   \\
    \hline
    mean    &46.6250    &3296.9059    &702.8232    &668.6643    &53.8293    &318.8873   \\
    \rowcolor{gray!10} std &26.7925    &1287.5402    &286.0818    &262.4843    &31.6477    &134.2972   \\
    min &1  &159    &34 &35  &0  &0   \\
    \rowcolor{gray!10} 25\% &28 &2407.5  &506    &482.5  &32 &227 \\
    50\%    &43    &3165  &676  &646   &49    &302   \\
    \rowcolor{gray!10} 75\% &61 &4081   &867.5  &828    &69 &396    \\
    max &307   &12855 &3859  &2631  &348   &1340  \\
    \hline
    \end{tabular}
\end{table}
%%%%%%%%%%%%%%%%%%%%%%%%%%%%%%%%----Sub-sub-Section----%%%%%%%%%%%%%%%%%%%%%%%%%%%%%%%%%%%%%%%%%%%%%
\subsubsection*{myPersonality}
myPersonality\footnote{\url{https://sites.google.com/michalkosinski.com/mypersonality}} is a collection of 250 anonymous Facebook users profile updates scored in Big Five by questioning users to answer them. Since 2018 creators, Stillwell and Kosinski stopped sharing and developing the dataset. There are some versions available on the internet that do not match on records but approximately it contained 9900 records. The myPersonality annotated in two forms, classification and regression, and table \ref{tbl:mypersonality_label_dist_binary} \& \ref{tbl:mypersonality_dataset_labels_dist} illustrate the distribution of status updates of myPersonality in each form.

\begin{table}[]
    \centering
    \caption{Distribution of labels in myPersonality classificational mode. \cite{Wang2020_Encoding}}
    \label{tbl:mypersonality_label_dist_binary}
    \begin{tabular}{cccccc}
        \hline
        \textbf{Label}  &\textbf{Opn}   &\textbf{Con}   &\textbf{Ext}   &\textbf{Agr}   &\textbf{Neu}    \\
        \hline
        Yes &176    &130    &96 &134    &99 \\
        No  &74 &120    &154    &116    &151    \\
        \hline
    \end{tabular}
\end{table}

\begin{table}[]
    \centering
    \caption{The distribution of myPersonality dataset labels in regresional mode. \cite{Darliansyah2019}}
    \label{tbl:mypersonality_dataset_labels_dist}
    \begin{tabular}{cccccc}
        \hline
        &\textbf{Opn}   &\textbf{Con}   &\textbf{Ext}   &\textbf{Agr}   &\textbf{Emo}  \\
        \hline
        Max    &5 &5    &5  &5  &4.75 \\
        \rowcolor{gray!10} Min &2.25	&1.45	&1.33	&1.65	&1.25 \\
        Avg &4.0786 &3.5229 &3.2921	&3.6003	&2.6272 \\
        \rowcolor{gray!10} Std  &0.5751 &0.7402 &0.8614 &0.5708 &0.7768    \\
        \hline
    \end{tabular}
\end{table}
%%%%%%%%%%%%%%%%%%%%%%%%%%%%%%%%----Sub-sub-Section----%%%%%%%%%%%%%%%%%%%%%%%%%%%%%%%%%%%%%%%%%%%%%
\subsubsection*{YouTube}
Youtube is the most popular video-sharing platform and has attracted many vloggers. YouTube dataset is a collection of 404 YouTube vloggers personality scores in big five. The dataset is recorded talking in front of webcam about a variety of topics and annotated using Amazon Mechanical Turk and the Ten-Item Personality Inventory (TIPI). As mentioned, YouTube dataset is a multimodal, video, speech, and transcript to text \cite{Biel2013,Biel2013text}. Based on the aim of the article, speech transcripts to text \cite{Biel2013text} is analyzed in this section. As the routine, distribution of traits has been shown in table \ref{tbl:youtube_dataset_labels} and textual elements details are shown in table \ref{tbl:youtube_dataset_elements}.

\begin{table}[]
    \centering
    \caption{The distribution of labels in YouTube dataset.}
    \label{tbl:youtube_dataset_labels}
    \begin{tabular}{cccccc}
        \hline
        &\textbf{Opn}   &\textbf{Con}   &\textbf{Ext}   &\textbf{Agr}   &\textbf{Emo}  \\
        \hline
        Max    &6.3 &6.2 &6.6	&6.5	&6.5 \\
        \rowcolor{gray!10} Min &2.4	&1.9	&2	&2	&2.2 \\
        Avg &4.6642	&4.4971	&4.6248	&4.6825	&4.7662 \\
        \rowcolor{gray!10} Std  &0.7154    &0.7701    &0.9771  &0.8793   &0.7790    \\
        \hline
    \end{tabular}
\end{table}

\begin{table}[]
    \centering
    \caption{The elements count of Texts in YouTube dataset.}
    \label{tbl:youtube_dataset_elements}
    \begin{tabular}{c|ccccccc}
        \hline
        &{\small \textbf{sentences}} &{\small \textbf{characters}} &{\small \textbf{spaces}} & {\small \textbf{words}} &{\small \textbf{duplicates}}   &{\small \textbf{punctuations}} &{\small \textbf{stop words}} \\
        \hline
        mean &	41.3490 &	3040.1905 &	595.0767 &	615.6014 &	95.5792 & 149.0519 & 	261.5272 	 \\
        \rowcolor{gray!10} std & 28.3491& 	2133.3875&	415.7767&	429.1693& 	51.3635&107.4581& 	189.7467 	 \\
        min &	2& 	155& 	24& 	25& 	5& 5& 5 \\
        \rowcolor{gray!10} 25\% &21 &1472   &287.75 &299.75 &55 &71.25  &120    \\
        50\%    &34 &2351.5 &466    &483.5  &85 &117.5  &201.5   \\
        \rowcolor{gray!10} 75\% &55 &3936.25    &767.50 &794    &123    &192.25 &345\\
        max  &158    &10814  &2031   &2099   &254    & 573 	&961    \\
        \hline
    \end{tabular}
\end{table}
%%%%%%%%%%%%%%%%%%%%%%%%%%%%%%%%----Sub-sub-Section----%%%%%%%%%%%%%%%%%%%%%%%%%%%%%%%%%%%%%%%%%%%%%
\subsubsection*{FriendPersona} \label{subsubsec:friendpersona_dataset}
The newest dataset introduced in the context of text-based APP is FriendPersona. This is developed on Friends TV Show Dataset\footnote{\url{https://github.com/emorynlp/character-mining}} \cite{Chen2016} and 711 conversation extracted from. FriendPersona annotated by three experts and to make it binary, split from the median. The dataset could be found in github\footnote{\url{https://github.com/emorynlp/personality-detection}}.% \hl{to be continued}
%%%%%%%%%%%%%%%%%%%%%%%%%%%%%%%%----Sub-sub-Section----%%%%%%%%%%%%%%%%%%%%%%%%%%%%%%%%%%%%%%%%%%%%%
\subsubsection*{Kaggle MBTI}
Kaggle MBTI\footnote{Available on \url{https://www.kaggle.com/datasnaek/mbti-type/}} has gathered 50 last posts through the PersonalityCafe forum in MBTI personality model. There are 8675 rows of data that each row represents a person. The dataset is started on cognitive functions by Carl Jung and finally personality tags done by Jungian Typology in MBTI.
%%%%%%%%%%%%%%%%%%%%%%%%%%%%%%%%%%----Sub-Section----%%%%%%%%%%%%%%%%%%%%%%%%%%%%%%%%%%%%%%%%%%%%%%%
\subsection{Results} \label{subsec:results}
In this part, firstly evaluation metrics used in APP introduced and then reported results of methods appeared in the division of datasets. Each following tables survey a dataset presented in the previous part.

\subsection{Evaluation metrics}
Precision, recall, accuracy, and F-measure are well-known Classification evaluation metrics are using in scientific reports. For calculating these measures, four concepts should be defined. TP, TN, FP, and FN are the notions and denotes to true positive, true negative, false positive, and false negative, respectively. These concepts make sense based on ground truth in confronting with the output of the system. There are other measurements for evaluating classification (Regression) methods such as RMSE (Root Mean Square Error), MAE (Mean Absolute Error), and Coefficient of Determination ($R^2$). Still, most articles prefered the binary classification measures and reported in four first measures. The following equations are the calculation of the metrics.

\begin{equation}
\label{eq:pr}
    Precision = \frac{\# \: system \: predicted \: true \: label}{\# \: items \: system \: predicted \: as \: true \: label} = \frac{TP}{TP \: + \: FP}
\end{equation}

\begin{equation}
\label{eq:re}
    Recall = \frac{\#  \: system  \: predicted  \: true  \: label}{\#  \: ground  \: truth  \: true  \: items} = \frac{TP}{TP  \: +  \: FN}
\end{equation}

\begin{equation}
\label{eq:acc}
    Accuracy = \frac{TP + TN}{TP +  TN +  FP +  FN}
\end{equation}

\begin{equation}
\label{eq:f1}
    F-measure = 2.\frac{Precision \times Recall}{Precision + Recall}
\end{equation}

F-1 score (Eq.\ref{eq:f1}) translates to harmonic mean of precision (eq.\ref{eq:pr}) and recall (eq.\ref{eq:re}) and that is why most studies report F1 score to make readers understand easier instead of reporting precision and recall.

\begin{equation}
\label{eq:rmse}
    RMSE = \sqrt{\frac{\sum_{t = 0}^{N - 1}{(y_t - \hat y_t)}^2}{N}}
\end{equation}

\begin{equation}
\label{eq:mae}
    MAE = \sum_{t = 0}^{N - 1} |(y_t - \hat y_t)|
\end{equation}

As respect of lack of text-based personality prediction studies, researchers reported the results on their aspect without much comparison, and this makes variety in results measurement units. RMSE (eq. \ref{eq:rmse}) and MAE (eq. \ref{eq:mae}) has been used in \cite{Carducci2018}, \cite{Xue2018}, however can not be compared cause of non comparable studies in same dataset.

\subsection{Discussion}
For evaluating and making results comparative and comprehensive, the results of each dataset are given in separate tables. The Essays is the most popular dataset in APP, and the number of results is more than the others. According to the classification essence of Essays and MyPersonality datasets, the evaluations are done by F-measure and Accuracy metrics. Augmentation of APP methods have been accelerated more rapidly with the advent of deep learning, and the increase in the quality of results is evident in Essays datasets more obviously. As it is shown in table \ref{tbl:results_essay}, the results of each method that powered the newer PLMs and novel deep learning method are achieved higher values compared with the previous method. The point that has to be considered is that some methods reported results insufficiently; thus, solely one evaluation metric is listed by personal opinion. Among the methods that do not use word embeddings and evaluated on the Essays dataset, \cite{Ramezani2020} is got better results. Since the MyPersonality dataset is not available by the creators anymore, it is not used in recent researches. However, \cite{Wang2020_Encoding} achieved the best results, as it is shown in table \ref{tbl:results_mypersonality}, which does not apply PLMs for embedding.

%\begin{landscape}
\begin{table}[]
%\begin{sidewaystable}
    \centering
    \caption{Experimental results on Essay dataset. The boldface values indicate the best values in each trait.}
    \label{tbl:results_essay}
    \tiny
    \begin{tabular}{c|ccccc:c|ccccc:c}
        \hline
        \multirow{2}{*}{\textbf{Method}}    &\multicolumn{6}{c}{\textbf{F-measure}} &\multicolumn{6}{|c}{\textbf{Accuracy}}   \\
        \cline{2-13}
            &\textbf{Opn}   &\textbf{Con}   &\textbf{Ext}   &\textbf{Agr}   &\textbf{Neu}   &\textbf{Avg}   &\textbf{Opn}   &\textbf{Con}   &\textbf{Ext}   &\textbf{Agr}   &\textbf{Neu}   &\textbf{Avg}   \\
        \hline
        %\citet{} (The large model)  &65.79  &\textbf{69.44} &66.01  &65.71  &65.32  &66.452 &60.67 &\textbf{67.04} &61.04  &\textbf{64.04} &\textbf{63.67}  &\textbf{63.292}  \\
        \citet{Xue2021} &\textbf{67.84}  &\textbf{63.46} &\textbf{71.50} &\textbf{71.92} &62.36  &67.416  &63.16 &57.49  &58.91  &57.49  &59.51   &59.312  \\
        \rowcolor{gray!10} \citet{Ren2021} &   &   &   &   &   &   &\textbf{80.35}  &\textbf{80.23}  &\textbf{79.94}  &\textbf{80.30}  &\textbf{80.14}  &\textbf{80.192} \\
        \citet{Ramezani2020} &57.37 &59.74  &65.80  &61.62  &60.69  &61.04  &56.30  &59.18  &64.25 &60.31 &61.14  &60.24 \\
        \rowcolor{gray!10} \citet{Wang2020_Encoding}   &67   &68 &67 &69    &\textbf{69}    &\textbf{68}    &64.80  &59.10  &60 &57.70  &63 &60.92  \\
        %\citet{Demerdash2020}   &   &   &   &   &   &   &63.30  &57.97  &58.85  &59.25  &59.88  &59.85  \\
        \citet{Jiang2020}   &   &   &   &   &   &   &65.86 &58.55  &60.62  &59.72  &61.04  &61.158 \\
        \rowcolor{gray!10} \citet{Mehta2020}   &   &   &   &   &    &   &64.6   &59.2   &60 &58.8    &60.5  &60.62  \\
        \citet{Kazameini2020}   &   &   &   &   &   &   &62.09  &57.84   &59.30    &56.52 &59.39    &59.028\\
        %\citet{Salminen2020}    &65.3  &54.3  &66.2  &60.3  &33.2    &55.86   &   &   &     &   &    &   \\
        \rowcolor{gray!10} \citet{Majumder2017} &   &   &   &   &   &   &62.68   &57.30   &58.09 &56.71  &59.38  &58.832 \\
        %\rowcolor{gray!10} \citet{YakutKilic2017}  &58 &36 &32 &30 &33 &37.8   &   &   &   &   &   &   \\
        %\citet{Wang2019} &70 &63 &62 &57 &63 &63   &67.58  &64.94  &64.34  &64.30  &65.53  &65.34  \\
        %\citet{Tighe2016}   &61.9    &56 &55.6   &55.7   &58.3   &57.7   &61.95  &56.04 &55.75  &57.54  &58.31   &57.918   \\
        \citet{Tighe2016}   &61.9    &56 &55.6   &55.7   &58.3   &57.7   &61.95  &56.04 &55.75  &57.54  &58.31   &57.918   \\
        \rowcolor{gray!10} \citet{Verhoeven2013}   &56 &54 &53 &50 &54 &53.4    &   &   &   &   &   &    \\
        \citet{Poria2013}   &66.1    &63.3   &63.4   &61.5   &63.7   &63.6   &   &   &   &   &   &   \\
        \rowcolor{gray!10} \citet{Mohammad2013}    &60.57  &56.46  &56.28  &53.9   &58.15  &57.072  &   &   &   &   &   & \\
        \hline
    \end{tabular}
\end{table}
%\end{sidewaystable}
%\end{landscape}

\begin{table}[]
    \centering
    \caption{Experimental results on MyPersonality dataset. The boldface values indicate the best values in each trait.}
    \label{tbl:results_mypersonality}
    \begin{tabular}{c|ccccc:c|ccccc:c}
        \hline
        \multirow{2}{*}{\textbf{Method}}  &\multicolumn{6}{c}{\textbf{F-measure}} &\multicolumn{6}{|c}{\textbf{Accuracy}}   \\
        \cline{2-13}
            &\textbf{Opn} &\textbf{Con} &\textbf{Ext} &\textbf{Agr} &\textbf{Neu}   &\textbf{Avg}   &\textbf{Opn} &\textbf{Con} &\textbf{Ext} &\textbf{Agr} &\textbf{Neu}   &\textbf{Avg} \\
        \hline
        \citet{Wang2020_Encoding}   &\textbf{76.8}  &75 &\textbf{85}    &\textbf{70}  &\textbf{79}  &\textbf{77.16}  &80 &76 &80 &68 &79 &\textbf{76.6}  \\
        %\citet{Drexel2019}  &52    &\textbf{78}   &63 &63 &59 &63 &   &   &   &   &   &   \\
        \rowcolor{gray!10} \citet{Zheng2019}   &65 &62 &71 &68 &70 &67.2    &   &   &   &   &   & \\
        \citet{Hassanein2019}    &   &   &   &   &   &65 &   &   &   &   &   &64 \\
        \rowcolor{gray!10} \citet{Farnadi2013}  &61 &54 &56 &45 &49  &53   &    &   &   &   &    &   \\
        \hline
    \end{tabular}
\end{table}

\begin{table}[]
    \centering
    \caption{Experimental results on YouTube dataset.}
    \label{tbl:results_youtube}
    \begin{tabular}{c|ccccc|ccccc}
        \hline
        \multirow{2}{*}{\textbf{Method}}  &\multicolumn{5}{c}{\textbf{F-measure}} &\multicolumn{5}{|c}{\textbf{RMSE}} \\
        \cline{2-11}
            &\textbf{Opn}   &\textbf{Con} &\textbf{Ext} &\textbf{Agr} &\textbf{Emo}   &\textbf{Opn}   &\textbf{Con}   &\textbf{Ext} &\textbf{Agr} &\textbf{Emo} \\
        \hline
        \citet{Xue2021} &78.57  &77.20  &82.35  &83.08   &79.55   \\
        %\citet{Salminen2020}    &68.6   &48.5   &71.9   &44.4   &40.3  &   &   &   &     &   \\
        %\rowcolor{gray!10} \citet{} (AlBert as the best result)  &   &   &   &   &   &0.6682    &0.955	&0.7944	&0.6089	&0.6486 \\
        \citet{Sun2020}  &   &   &   &   &   &0.6813    &0.6898    &0.8897  &0.7743 &0.6872 \\
        \hline
    \end{tabular}
\end{table}

As mentioned in Sec \ref{subsubsec:friendpersona_dataset}, FriedndPersona is the most recent APP dataset that was only introduced for the proposed method on \cite{Jiang2020} and are shown in table \ref{tbl:results_friendpersona}. But in comparison with reported results on Essays (table \ref{tbl:results_essay}) the evaluations are acceptable because the proposed model achieved in range of 60\% (63.01\% on FriendPersona and 61.158\% on Essays).% \hl{To be continued. Based on distribution of dataset, results will be expanded.}

\begin{table}[]
    \centering
    \caption{Experimental results on FriendPersona dataset.}
    \label{tbl:results_friendpersona}
    \begin{tabular}{c|ccccc|ccccc}
        \hline
        \multirow{2}{*}{\textbf{Method}}  & \multicolumn{5}{c}{\textbf{F-measure}} & \multicolumn{5}{|c}{\textbf{Accuracy}}   \\
        \cline{2-11}
            &\textbf{Opn}   &\textbf{Con}   &\textbf{Ext}   &\textbf{Agr}   &\textbf{Neu} &\textbf{Opn}   &\textbf{Con}   &\textbf{Ext}   &\textbf{Agr}   &\textbf{Neu} \\
        \hline
        \citet{Jiang2020}    &   &   &   &   &   &68.47 &56.78   &64.21 &65.58  &60.06 \\
        \hline
    \end{tabular}
\end{table}

The evaluations on the Kaggle dataset are shown in table \ref{tbl:results_kaggle} and done using f-measure and accuracy metrics. The best-performed method is \cite{Khan2020} that deployed XGboost ensemble algorithm to gain high values on both metrics. To the rest methods, three reported accuracies and one reported f-measure only that does not have any subscription, so methods with the same reported metrics should be compared.

\begin{table}[]
    \centering
    \caption{Experimental results on Kaggle MBTI dataset. The boldface values indicate the best values in each trait.}
    \label{tbl:results_kaggle}
    \begin{tabular}{c|cccc|cccc}
        \hline
        \multirow{2}{*}{\textbf{Method}}  &\multicolumn{4}{c}{\textbf{F-measure}} &\multicolumn{4}{|c}{\textbf{Accuracy}}   \\
        \cline{2-9}
            &\textbf{I-E}   &\textbf{S-N}   &\textbf{F-T}   &\textbf{J-P}   &\textbf{I-E} &\textbf{S-N}   &\textbf{F-T}   &\textbf{J-P}   \\
        \hline
        \citet{Khan2020}    &\textbf{98.56}  &\textbf{99.75}  &\textbf{94.72}  &\textbf{96.19}  &\textbf{99.37}  &\textbf{99.92}  &\textbf{94.55}  &\textbf{95.53} \\
        \rowcolor{gray!10} \citet{Yang2021}{}  &66.08  &69.10  &79.19  &67.50  &   &   &   &   \\
        \citet{Amirhosseini2020} &   &   &   &   &78.17  &86.06  &71.78  &65.70\\
        \rowcolor{gray!10} \citet{Mehta2020}   &   &   &   &   &78.8   &86.3   &76.1   &67.2  \\
        %\citet{Rayne2017} &   &   &   &  &54.0  &52.9   &57.8   &52.9   \\
        \hline
    \end{tabular}
\end{table}

In the end, as it can be concluded from the tables, deployment of PLMs for words and documents embeddings in couple with deep neural network models makes a significant improvement in textual-based APPs. However, somehow, it seems that the hybrid and ensemble models (especially PLM-free) are achieving better results in their own class and could be a progressive area of research. Also, the APP research area suffers from a lack of golden standard datasets and evaluations. As the final recommendation for future work, introducing novel datasets labelled on more than one personality trait model, with more samples and different lengths of documents, can make the APP research area more attractive and competitive.
%%%%%%%%%%%%%%%%%%%%%%%%%%%%%%%%%%%%----Section----%%%%%%%%%%%%%%%%%%%%%%%%%%%%%%%%%%%%%%%%%%%%%%%%%
\section{Conclusion} \label{sec:conclusion}
Automatic personality prediction (perception) (APP) system provides an opportunity to predict personality traits based on human behaviour manifestations, especially in texts in this review. This paper is reviewed text-based APP methods since 2010 and reported results from five well-known benchmark datasets. To do so, articles have been categorized into three PLM-free approaches, PLM-based methods, and multimodal techniques. Also, the framework of overviewed methods has been collected. The aim of this review is to give a general overview of the steps of getting meliorate of APPs to researchers in this field. This review solely looks at existing personality trait identification methods from a computational standpoint and ignores psychological studies on personality trait recognition. Finally, we introduced several open issues that outline promising directions for future research on text-based automatic personality prediction:
\begin{itemize}
    \item As psychologists recognise that individuals personality traits change through changes in their speaking/writing, similarly, when the appeared words and expressions in the given input text change, the machine assigns different labels for traits, So the dataset, in fact, is the basis of the APP system, because a computer is learning from the dataset. All things considered, generating new datasets labelled by psychologists through essays and other features would be an excellent benchmark for evaluating personality prediction methods. Meaningfully, generate a dominant dataset and benchmark, that the results are based on convincing a psychologist and learning a computer.
    \item Making available datasets in several personality trait models simultaneously. This would make results more comprehensive and represent APP systems output psychologistic.
    \item Generating datasets in low-resource languages. Since languages characterise different features of people, different languages would make various personality-related features through texts.
\end{itemize}
%%%%%%%%%%%%%%%%%%%%%%%%%%%%%%%%%%%%----Section----%%%%%%%%%%%%%%%%%%%%%%%%%%%%%%%%%%%%%%%%%%%%%%%%%

\begin{acknowledgements}
This project is supported by a research grant from the University of Tabriz (number S/806).
\end{acknowledgements}

%%%%%%%%%%%%%%%%%%%%%%%%%%%%%%%%%%%----Section----%%%%%%%%%%%%%%%%%%%%%%%%%%%%%%%%%%%%%%%%%%%%%%%%%
\section*{Declarations}
\subsection*{Funding}
This project is supported by a research grant of the University of Tabriz (number S/806).

\subsection*{Conflict of interest}
The authors declare that they have no conflict of interest.

%\subsection{Authors' contributions}
%Ali-Reza Feizi-Derakhshi had the idea for the article and performed data analysis and first draft, Majid Ramezani, Meysam Asgari-Chenaglu, and Narjes Nikzad-Khasmakhi performed the literature search, and the all authors critically revised and edited the work.

\subsection{Data availability}
Not applicable.

\subsection{Ethics approval and consent to participate}
Not applicable.

\subsection{Consent for publication}
Not applicable.

\bibliographystyle{unsrtnat}
\bibliography{Bibliography.bib}   % name your BibTeX data base

%%%%%%%%%%%%%%%%%%%%%%%%%%%%%%%%%%%%----Section----%%%%%%%%%%%%%%%%%%%%%%%%%%%%%%%%%%%%%%%%%%%%%%%%%
\newpage
\appendix
\section{Appendix} \label{sec:appendix}

\begin{figure}
    \centering
    \includegraphics[width=0.7\textwidth]{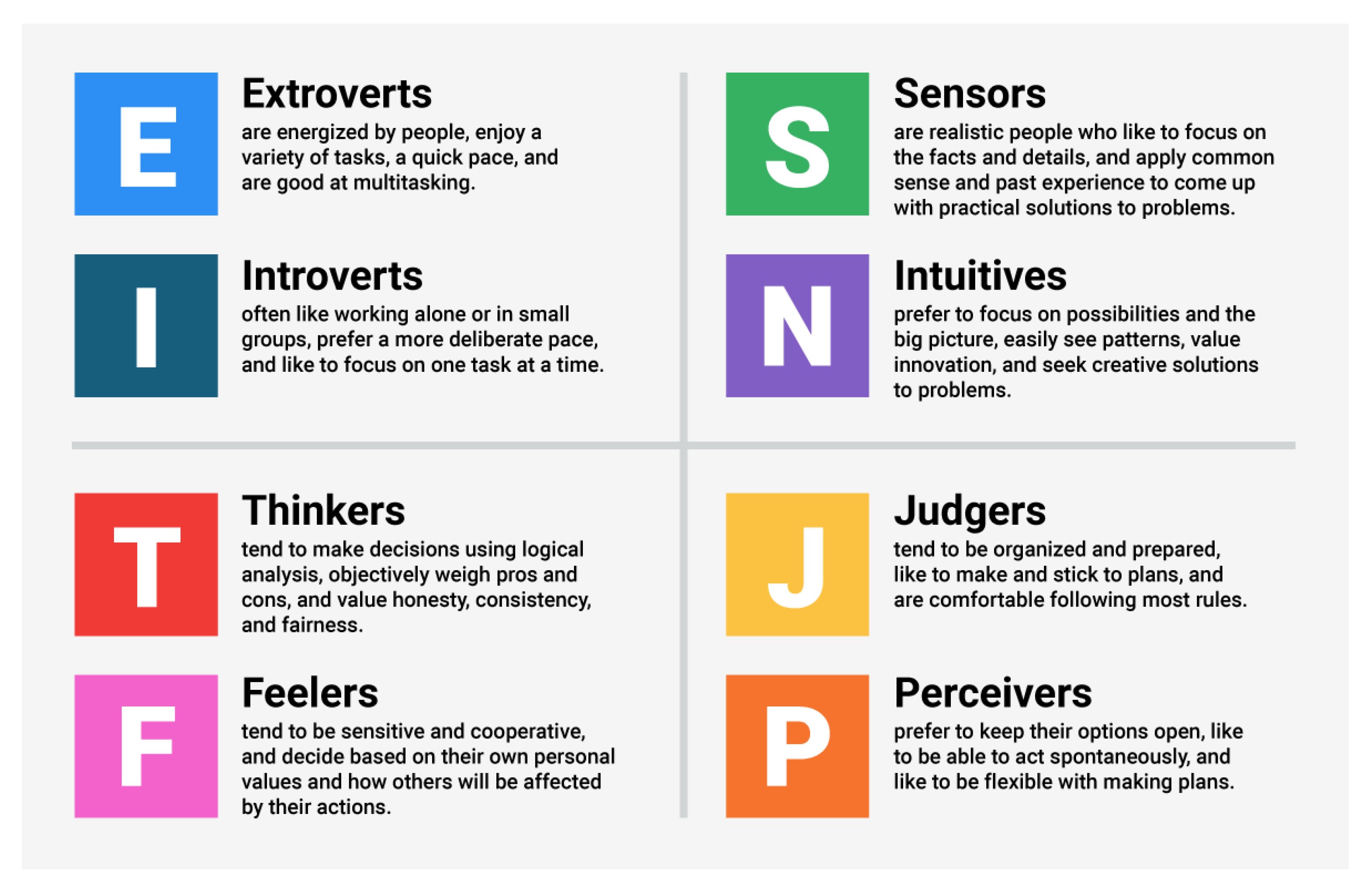}
    \caption{An overview of MBTI personality traits model \cite{Amirhosseini2020}.}
    \label{fig:mbti-overview}
\end{figure}

\begin{table}
	\scriptsize
	{
		\begin{center}
			\caption{An overview of Cattell's 16 \cite{cattel16-Wikipedia}.}
			\label{tbl:cattel16-overview}
			\fontsize{7}{8}
			\begin{adjustbox}{max width=\textwidth}
				\begin{tabular}{@{}p{5cm}@{}p{3.6cm}@{}p{5cm}@{}} % <-- Alignments: 1st column left, 2nd middle and 3rd right, with vertical lines in between
					
					\hline
					\hline
					\cellcolor{gray!15}	\textbf{Description of \textit{LOW} values \textcolor{gray} {-}} 	\textcolor{gray} {$\infty$ $\longleftarrow$}  	&	 \cellcolor{gray!15}	\textbf{Personality trait} \centering	&  		\cellcolor{gray!15}	\textcolor{gray} {$\longrightarrow$ +$\infty$}	 \textbf{ Description of \textit{HIGH} values}\\
					\hline
					\hline
					%--------   Warmth ---------
					& \textbf{Warmth (A)} \centering & 	
					\\
					\hline
					
					\begin{minipage}[t]{\columnwidth}
					\tiny
					\begin{tabular}{p{2.5cm}p{2.5cm}}
							•	Impersonal  &•	Distant \\
							•	Cool      &•	Reserved    \\
							•	Detached    &•	Formal  \\
							•	Aloof   &
						
					\end{tabular}
					\end{minipage}
					&   & 
					\tiny
					\begin{tabular}{p{2.5cm}p{2.5cm}}
							•	Warm    &•	Outgoing    \\
							•	Attentive to others &•	Kindly  \\
							•	Easygoing   &•	Participating    \\
							•	Likes people    &
					\end{tabular}
					
					\\
					\hline
					
					%--------   Reasoning ---------
					& 	\textbf{Reasoning (B)} 	\centering	 & 
					
					\\
					\hline
					
					\begin{minipage}[t]{\columnwidth}
					\tiny
					\begin{tabular}{p{2.5cm}p{2.5cm}}
						%		% In order to Wrap text and evoide line exceedings to next cells
							•	Concrete-thinking   &•	less intelligent    \\
							•	lower general mental capacity   &•	unable to handle abstract problems
					\end{tabular}
					\end{minipage}
					
					& &
					
					\begin{minipage}[t]{\columnwidth}
					\tiny
					\begin{tabular}{p{2.5cm}p{2.5cm}}
							•	Abstract-thinking   &•	more intelligent    \\
							•	higher general mental capacity  &•	bright  \\
							•	fast-learner    &
					\end{tabular}
					\end{minipage}
					\\
					
					\hline
					
					%--------   Emotional Stability (C) ---------
					&	 \textbf{Emotional Stability (C)} 		\centering	 & 
					\\
					\hline
					
					\tiny		%  To minimize the font size
					\begin{tabular}{p{2.5cm}p{2.5cm}}
							•	Reactive emotionally    &•	changeable  \\
							•	affected by feelings    &•	emotionally less stable \\
							•	easily upset    &
					\end{tabular}
					
					& & 
					
					\tiny		%  To minimize the font size
					\begin{tabular}{p{2.5cm}p{2.5cm}}
							•	Emotionally stable  &•	adaptive    \\
							•	faces reality calmly    &•	mature
					\end{tabular}
					
					\\
					
					\hline
					
					%--------   Dominance (E) ---------
					&	 \textbf{Dominance (E)}  \centering		& 
					\\
					\hline
					
					\tiny		%  To minimize the font size
					\begin{tabular}{p{2.5cm}p{2.5cm}}
							•	Deferential &•	cooperative \\
							•	avoids conflict &•	submissive  \\
							•	humble  &•	obedient    \\
							•	easily led  &•	docile  \\
							•	accommodating   &   
					\end{tabular}
					
					& & 
					
					\tiny		%  To minimize the font size
					\begin{tabular}{p{2.5cm}p{2.5cm}}
							•	Dominant    &•	forceful    \\
							•	assertive   &•	aggressive  \\
							•	competitive &•	stubborn    \\
							•	bossy   &
					\end{tabular}
					
					\\
					
					\hline
					
					%--------   Liveliness (F) ---------
					& 		\textbf{Liveliness (F)} 		\centering		 & 
					\\
					\hline
					
					\tiny		%  To minimize the font size
					\begin{tabular}{p{2.5cm}p{2.5cm}}
							•	Serious &•	restrained  \\
							•	prudent &•	taciturn    \\
							•	introspective   &•	silent
					\end{tabular}
					
					& & 
					
					\tiny		%  To minimize the font size
					\begin{tabular}{p{2.5cm}p{2.5cm}}
							•	Lively  &•	animated    \\
							•	spontaneous &•	enthusiastic    \\
							•	happy-go-lucky  &•	cheerful    \\
							•	expressive  &•	impulsive
						\end{tabular}
					
					\\
					\hline
					
					%--------   Rule-Consciousness (G) ---------
					& 		\textbf{Rule-Consciousness (G)} 		\centering		 & 
					\\
					\hline
					
					\tiny		%  To minimize the font size
					\begin{tabular}{p{2.5cm}p{2.5cm}}
							•	Expedient   &•	nonconforming   \\
							•	disregards rules    &•	self-indulgent
					\end{tabular}
					
					& & 
					
					\tiny		%  To minimize the font size
					\begin{tabular}{p{2.5cm}p{2.5cm}}
    							•	Rule-conscious  &•	dutiful \\
    							•	conscientious   &•	conforming  \\
    							•	moralistic  &•	staid   \\
    							•	rule-bound  &
						\end{tabular}
					
					\\
					\hline
					
					%--------   Social Boldness (H) ---------
					& 		\textbf{Social Boldness (H)} 		\centering		 & 
					\\
					\hline
					
					\tiny		%  To minimize the font size
					\begin{tabular}{p{2.5cm}p{2.5cm}}
							•	Shy &•	threat-sensitive    \\
							•	timid   &•	hesitant    \\
							•	intimidated &
					\end{tabular}
					
					& & 
					
					\tiny		%  To minimize the font size
					\begin{tabular}{p{2.5cm}p{2.5cm}}
							•	Socially bold   &•	venturesome \\
							•	thick-skinned   &•	uninhibited
						\end{tabular}
					
					\\
					\hline
					
					%--------  Sensitivity (I) ---------
					& 		\textbf{Sensitivity (I)} 		\centering		 & 
					\\
					\hline
					
					\tiny		%  To minimize the font size
					\begin{tabular}{p{2.5cm}p{2.5cm}}
							•	Utilitarian &•	objective   \\
							•	unsentimental   &•	tough-minded    \\
							•	self-reliant    &•	no-nonsense \\
							•	rough   &
					\end{tabular}
					
					& & 
					
					\tiny		%  To minimize the font size
					\begin{tabular}{p{2.5cm}p{2.5cm}}
							•	Sensitive   &•	aesthetic   \\
							•	sentimental &•	tender-minded   \\
							•	intuitive   &•	refined \\
						\end{tabular}
					
					\\
					\hline
					
					%--------   Vigilance (L) ---------
					& 		\textbf{Vigilance (L)} 		\centering		 & 
					\\
					\hline
					
					\tiny		%  To minimize the font size
					\begin{tabular}{p{2.5cm}p{2.5cm}}
							•	Trusting    &•	unsuspecting    \\
							•	accepting   &•	unconditional   \\
							•	easy    &
					\end{tabular}
					
					& & 
					
					\tiny		%  To minimize the font size
					\begin{tabular}{p{2.5cm}p{2.5cm}}
							•	Vigilant    &•	suspicious  \\
							•	skeptical   &•	distrustful \\
							•	oppositional    &
						\end{tabular}
					
					\\
					\hline
					
					%--------   Abstractedness (M) ---------
					& 		\textbf{Abstractedness (M)} 		\centering		 & 
					\\
					\hline
					
					\tiny		%  To minimize the font size
					\begin{tabular}{p{2.5cm}p{2.5cm}}
							•	Grounded    &•	practical   \\
							•	prosaic &•	solution oriented   \\
							•	steady  &•	conventional
					\end{tabular}
					
					& & 
					
					\tiny		%  To minimize the font size
					\begin{tabular}{p{2.5cm}p{2.5cm}}
							•	Abstract    &•	imaginative \\
							•	absentminded    &•	impractical \\
							•	absorbed in ideas   &
						\end{tabular}
					
					\\
					\hline
					
					%--------   Privateness (N) ---------
					& 		\textbf{Privateness (N)} 		\centering		 & 
					\\
					\hline
					
					\tiny		%  To minimize the font size
					\begin{tabular}{p{2.5cm}p{2.5cm}}
							•	Forthright  &•	genuine \\
							•	artless &•	open    \\
							•	guileless   &•	naive   \\
							•	unpretentious   &•	involved
					\end{tabular}
					
					& & 
					
					\tiny		%  To minimize the font size
					\begin{tabular}{p{2.5cm}p{2.5cm}}
							•	Private &•	discreet    \\
							•	nondisclosing   &•	shrewd  \\
							•	polished    &•	worldly \\
							•	astute  &•	diplomatic
						\end{tabular}
					
					\\
					\hline
					
					%--------   Apprehension (O) ---------
					& 		\textbf{Apprehension (O)} 		\centering		 & 
					\\
					\hline
					
					\tiny		%  To minimize the font size
					\begin{tabular}{p{2.5cm}p{2.5cm}}
							•	Self-assured    &•	unworried   \\
							•	complacent  &•	secure  \\
							•	free of guilt   &•	confident   \\
							•	self-satisfied  &
					\end{tabular}
					
					& & 
					
					\tiny		%  To minimize the font size
					\begin{tabular}{p{2.5cm}p{2.5cm}}
							•	Apprehensive    &•	self-doubting   \\
							•	worried &•	guilt-prone \\
							•	insecure    &•	worrying    \\
							•	self-blaming    &
						\end{tabular}
					
					\\
					\hline
					
					%--------   Openness to Change (Q1) ---------
					& 		\textbf{Openness to Change (Q1)} 		\centering		 & 
					\\
					\hline
					
					\tiny		%  To minimize the font size
					\begin{tabular}{p{2.5cm}p{2.5cm}}
							•	Traditional &•	attached to familiar    \\
							•	conservative    &•	respecting traditional ideas
					\end{tabular}
					
					& & 
					
					\tiny		%  To minimize the font size
					\begin{tabular}{p{2.5cm}p{2.5cm}}
							•	Open to change  &•	experimental    \\
							•	liberal &•	analytical  \\
							•	critical    &•	freethinking    \\
							•	flexibility &
						\end{tabular}
					
					\\
					\hline
					
					%--------   Self-Reliance (Q2) ---------
					& 		\textbf{Self-Reliance (Q2)} 		\centering		 & 
					\\
					\hline
					
					\tiny		%  To minimize the font size
					\begin{tabular}{p{2.5cm}p{2.5cm}}
							•	Group-oriented  &•	affiliative \\
							•	a joiner and follower dependent
					\end{tabular}
					
					& & 
					
					\tiny		%  To minimize the font size
					\begin{tabular}{p{2.5cm}p{2.5cm}}
							•	Self-reliant    &•	solitary    \\
							•	resourceful &•	individualistic \\
							•	self-sufficient &
						\end{tabular}
					
					\\
					\hline
					
					%--------   Perfectionism (Q3) ---------
					& 		\textbf{Perfectionism (Q3)} 		\centering		 & 
					\\
					\hline
					
					\tiny		%  To minimize the font size
					\begin{tabular}{p{2.5cm}p{2.5cm}}
							•	Tolerates disorder  &•	unexacting  \\
							•	flexible    &•	undisciplined   \\
							•	lax &•	self-conflict   \\
							•	impulsive   &•	careless of social rules    \\
							•	uncontrolled    &
					\end{tabular}
					
					& & 
					
					\tiny		%  To minimize the font size
					\begin{tabular}{p{2.5cm}p{2.5cm}}
							•	Perfectionistic &•	organized   \\
							•	compulsive  &•	self-disciplined    \\
							•	socially precise    &•  control \\
							•	exacting will power &•	self-sentimental
						\end{tabular}
					
					\\
					\hline
					
					%--------   Tension (Q4) ---------
					& 		\textbf{Tension (Q4)} 		\centering		 & 
					\\
					\hline
					
					\tiny		%  To minimize the font size
					\begin{tabular}{p{2.5cm}p{2.5cm}}
							•	Relaxed &•	placid  \\
							•	tranquil    &•	torpid  \\
							•	patient &•	composed low drive
					\end{tabular}
					
					& & 
					
					\tiny		%  To minimize the font size
					\begin{tabular}{p{2.5cm}p{2.5cm}}
							•	Tense   &•	high-energy \\
							•	impatient   &•	driven  \\
							•	frustrated  &•	over-wrought    \\
							•	time-driven &
						\end{tabular}
					
					\\
					\hline
				\end{tabular}
			\end{adjustbox}
		\end{center}
	}
\end{table}

\begin{figure}
    \centering
    \includegraphics[width=0.6\textwidth]{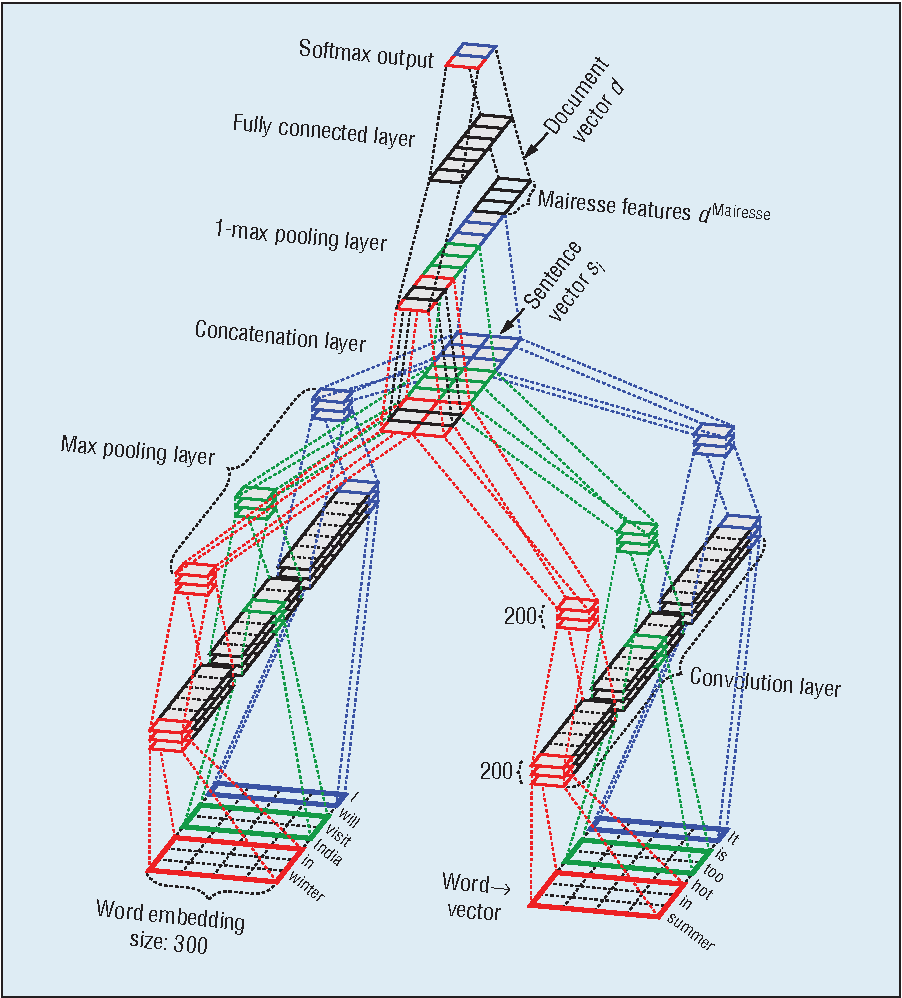}
    \caption{The architecture of method proposed in \cite{Majumder2017}. The input layer corresponds to the sequence of input sentences (only two are shown). The next two layers include three parts, corresponding to trigrams, bigrams, and unigrams. The dotted lines delimit the area in a previous layer to which a neuron of the next layer is connected—for example, the bottom-right rectangle shows the area comprising three-word vectors connected with a trigram neuron.}
    \label{fig:mjmuder-2017}
\end{figure}

\begin{figure}
    \centering
    \includegraphics[width=0.4\textwidth]{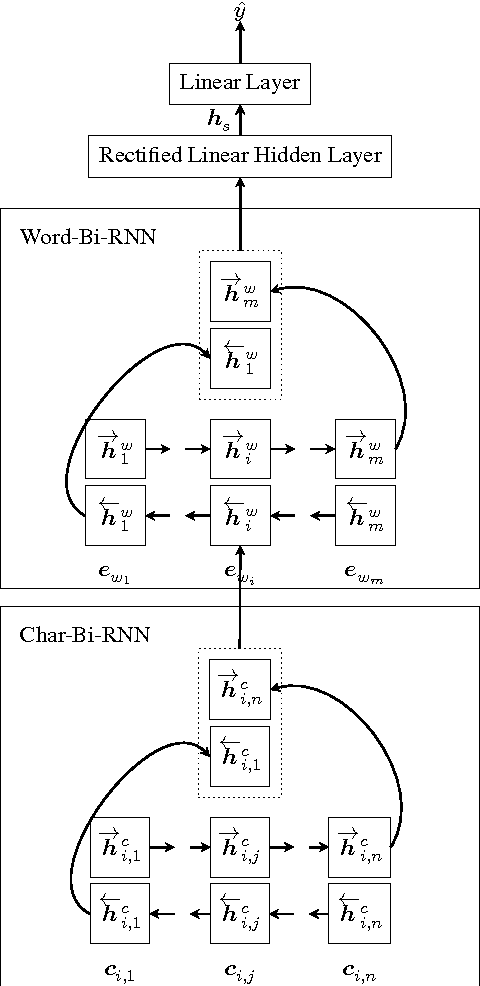}
    \caption{The architecture of proposed model for representation of texts called C2W2S4PT (Character to Word to Sentence for Personality Traits). Dotted boxes indicate concatenation. \cite{FLiu2016,Liu2017}}
    \label{fig:FLiu2016}
\end{figure}

% \begin{figure}
%     \centering
%     \includegraphics[width=0.6\textwidth]{Yakut-2017.jpg}
%     \caption{\cite{YakutKilic2017}}
%     \label{fig:}
% \end{figure}

\begin{figure}
    \centering
    \includegraphics[width=0.8\textwidth]{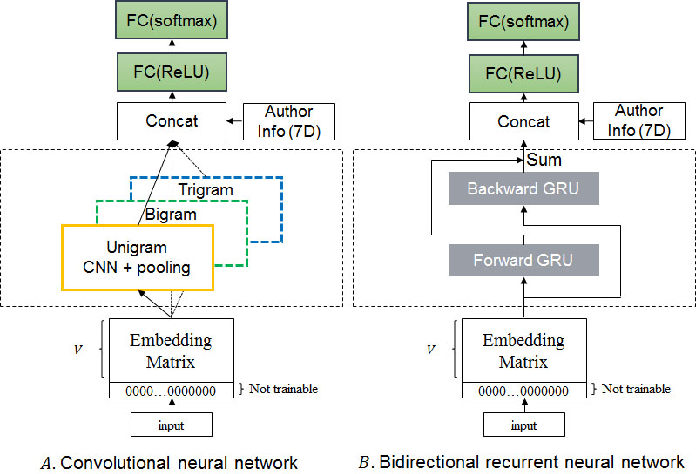}
    \caption{The proposed architectures in\cite{Yu2017}.}
    \label{fig:Yu2017}
\end{figure}

\begin{figure}
    \begin{subfigure}{.45\textwidth}
        \includegraphics[width=\linewidth]{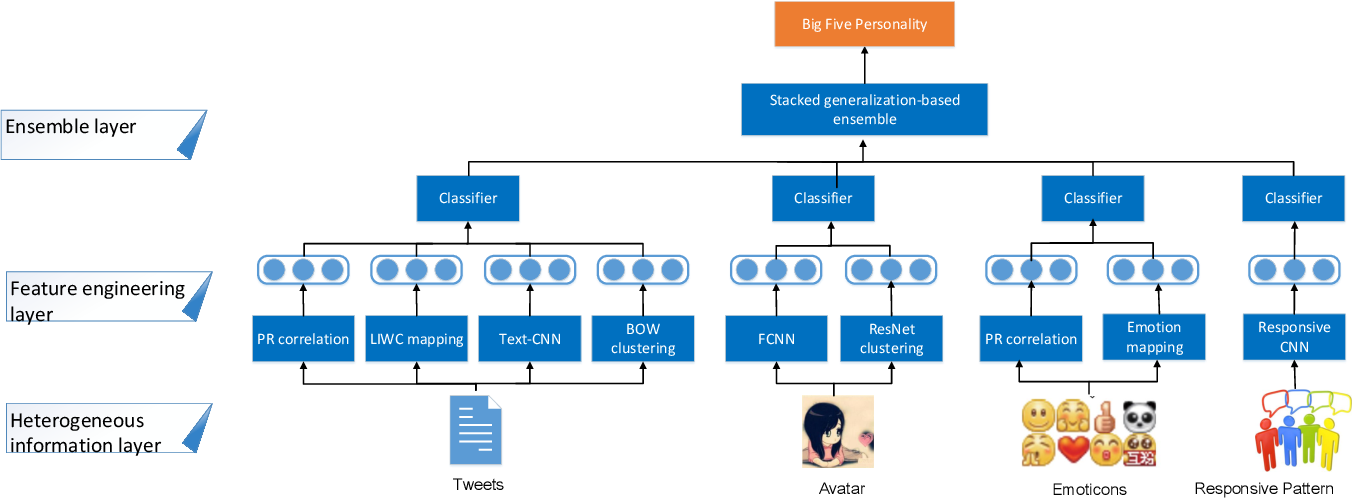}
        \caption{Framework of Heterogeneous Information Ensemble (HIE)}
    \end{subfigure}
    \begin{subfigure}{.45\textwidth}
        \includegraphics[width=\linewidth]{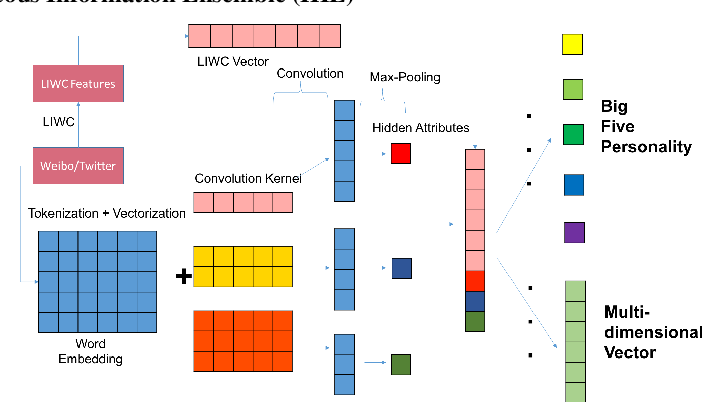}
        \caption{The structure of Text-CNN.}
    \end{subfigure}
    \caption{The structure of proposed algorithm called HIE and text representation model. \cite{Wei2017}}
    \label{fig:Wei2017}
\end{figure}

% \begin{figure}
%     \centering
%     \includegraphics[width=0.6\textwidth]{Xue-2018.png}
%     \caption{\cite{Xue2018}}
%     \label{fig:}
% \end{figure}

\begin{figure}
    \centering
    \includegraphics[width=\textwidth]{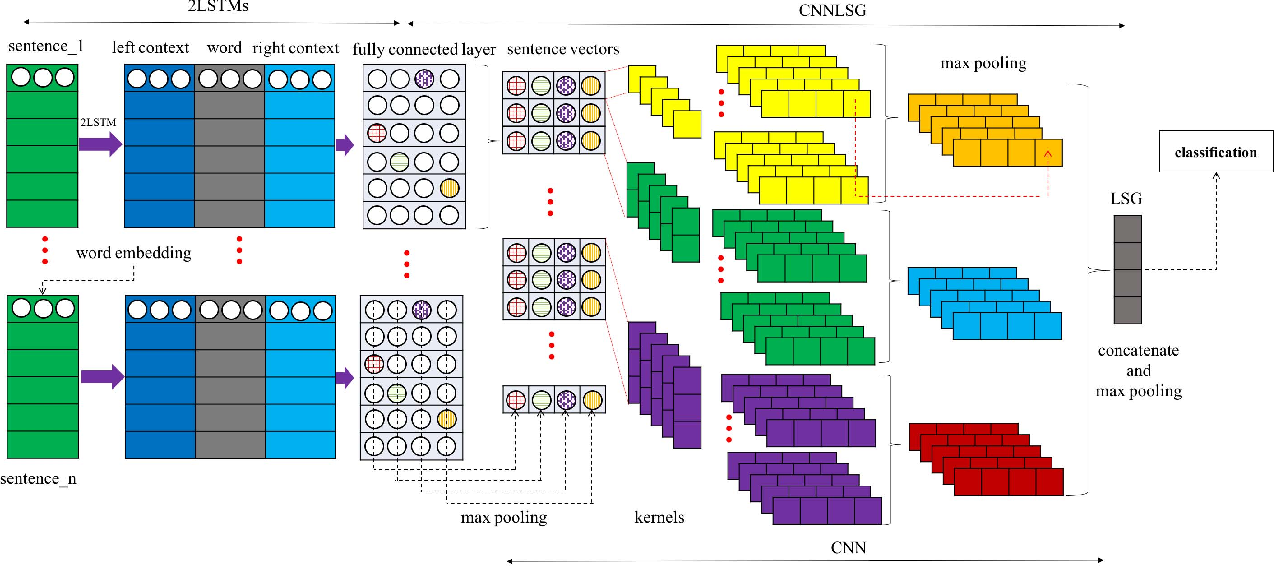}
    \caption{The architecture of proposed method in \cite{Sun2018} called 2CLSTM.}
    \label{fig:2CLSTM_Sun2018}
\end{figure}

% \begin{figure}
%     \centering
%     \includegraphics[width=\textwidth]{Kumar-2019.png}
%     \caption{\cite{Kumar2019}}
%     \label{fig:}
% \end{figure}

% \begin{figure}
%     \centering
%     \includegraphics[width=\textwidth]{Akrami-2019.png}
%     \caption{\cite{Akrami2019}}
%     \label{fig:}
% \end{figure}

\begin{figure}
    \centering
    \includegraphics[width=\textwidth]{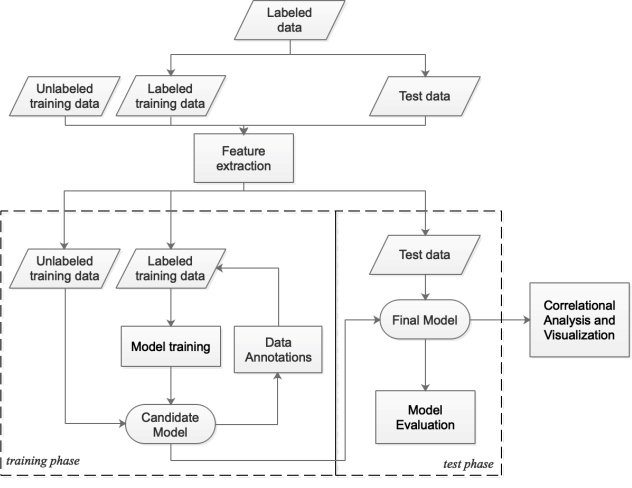}
    \caption{Pseudo Multi-view Co-training (PMC) based framework for persinality prediction \cite{Zheng2019}.}
    \label{fig:Zheng2019}
\end{figure}

\begin{figure}
    \centering
    \includegraphics[width=\textwidth]{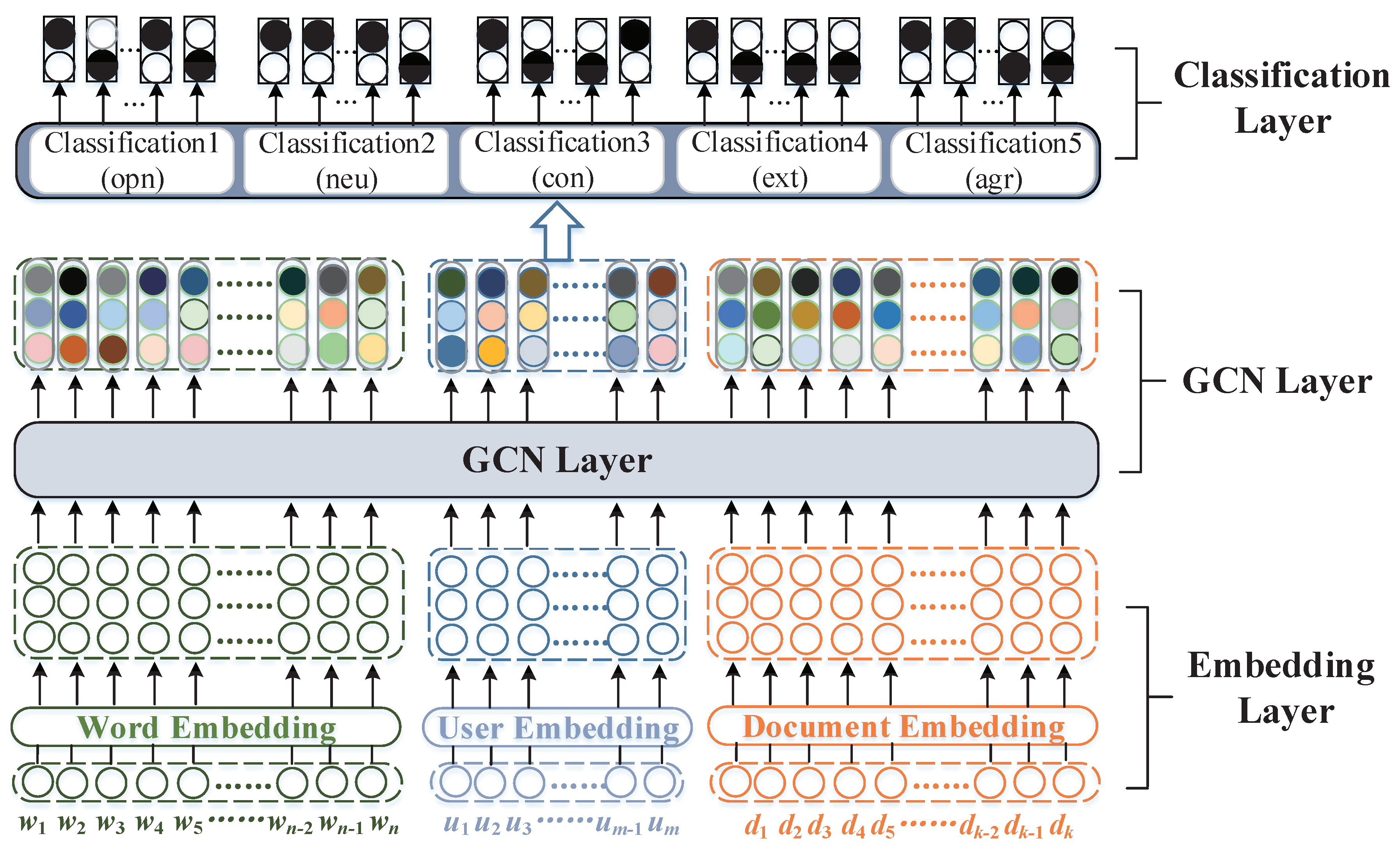}
    \caption{Three layer architecture overview of proposed personality GCN \cite{Wang2020_Encoding}.}
    \label{fig:Wang2020_Encoding}
\end{figure}

\begin{figure}
    \centering
    \includegraphics[width=\textwidth]{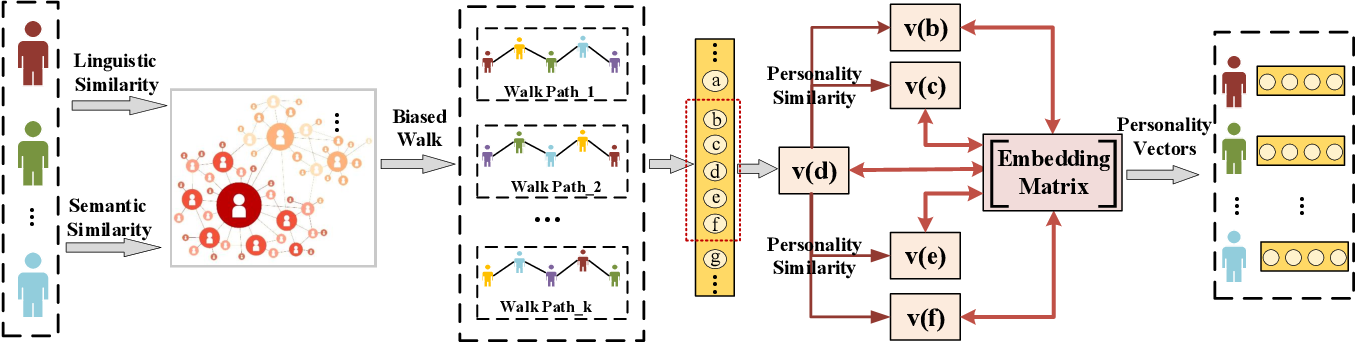}
    \caption{Overall framework of Personality2Vec \cite{Guan2020}.}
    \label{fig:Guan2020}
\end{figure}

\begin{figure}
    \centering
    \includegraphics[width=\textwidth]{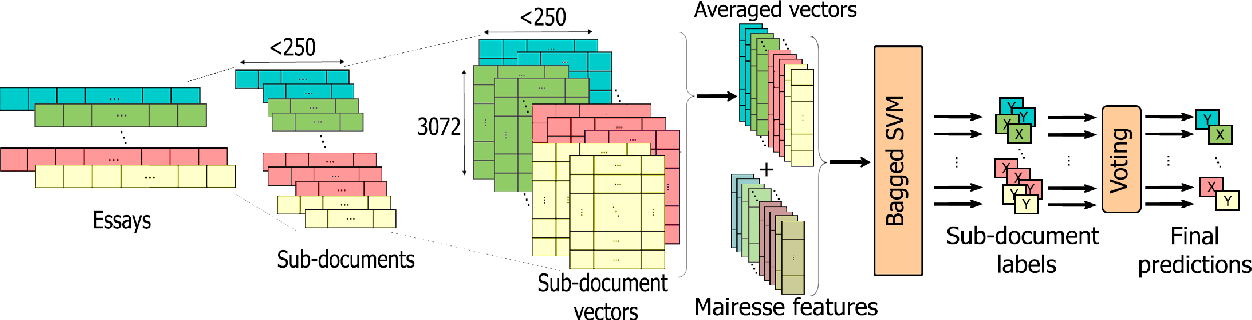}
    \caption{The architecture of Bagged SVM over BERT Word Embedding technique \cite{Kazameini2020}.}
    \label{fig:Kazameini2020}
\end{figure}

\begin{figure}
    \centering
    \includegraphics[width=0.8\textwidth]{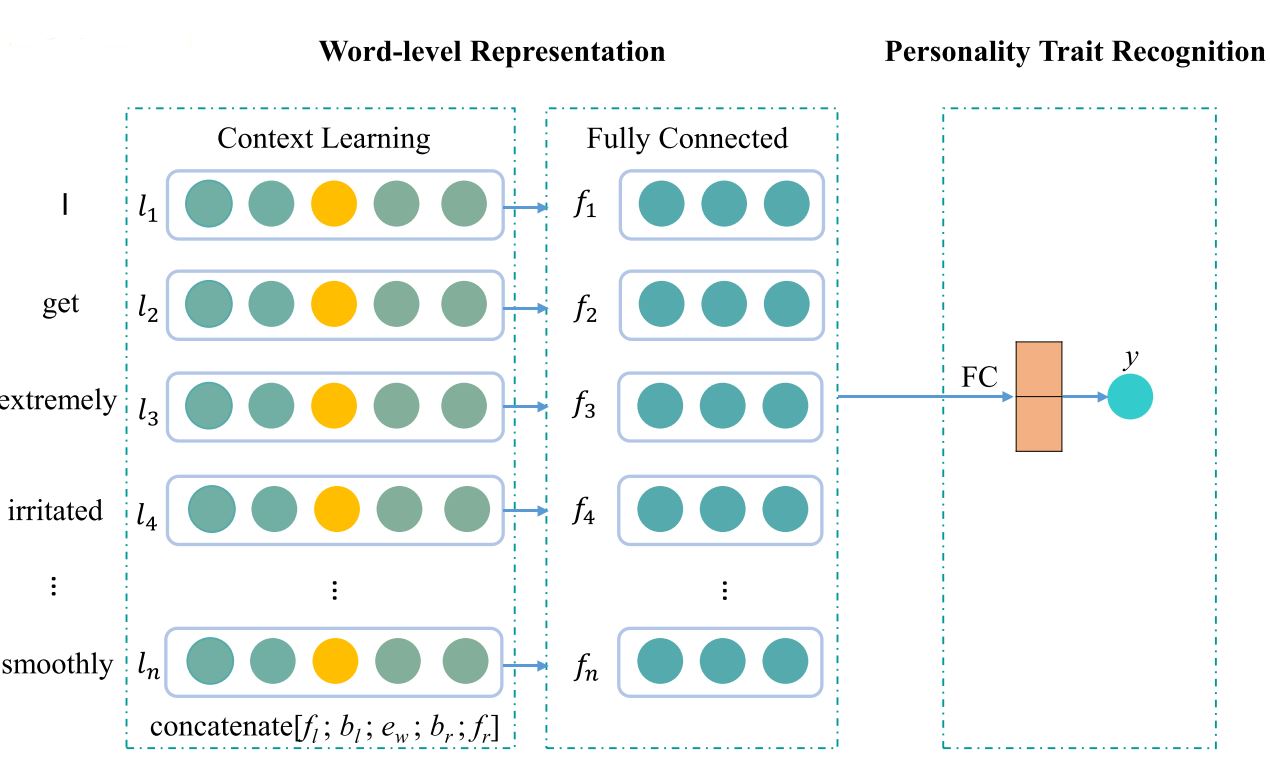}
    \caption{Illustration of SEPRNN (semantic-enhanced personality recognition neural network) \cite{Xue2021}.}
    \label{fig:Xue2021}
\end{figure}

\begin{figure}
    \centering
    \begin{subfigure}{0.64\textwidth}
        \centering
        \includegraphics[width=\textwidth]{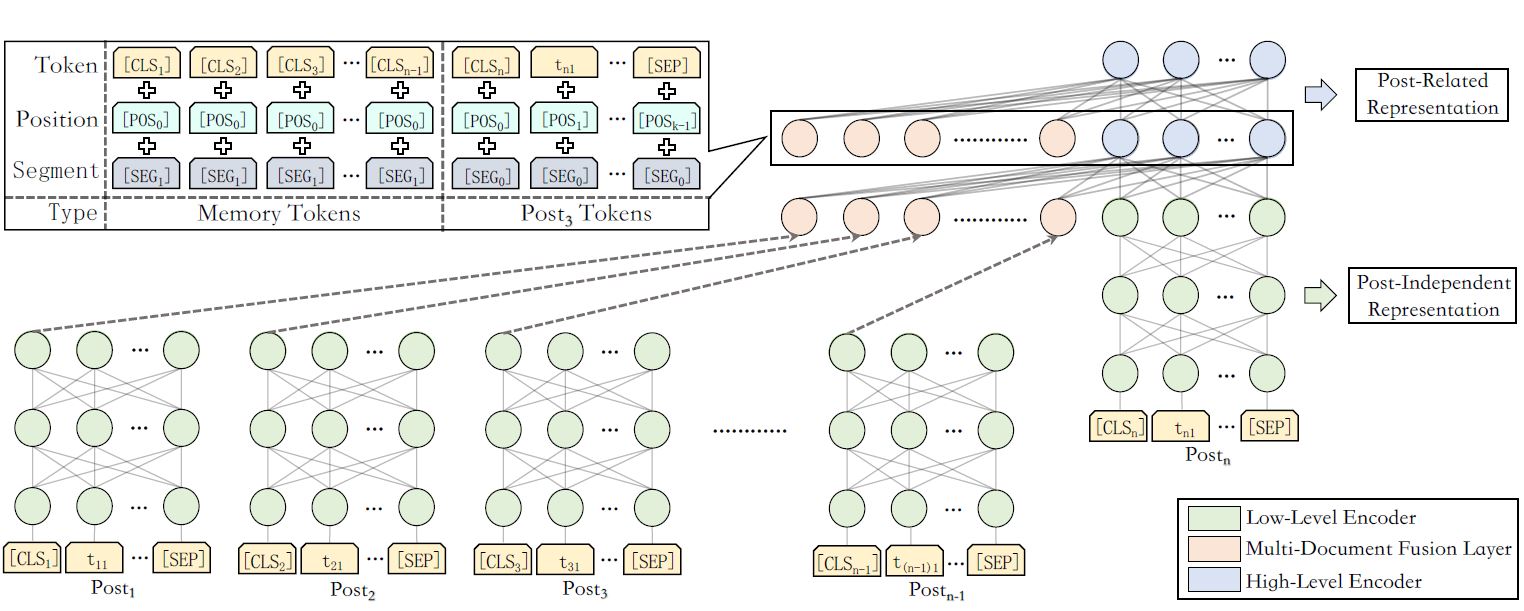}
        \caption{Overview of Transformer-MD.}
    \end{subfigure}
    \begin{subfigure}{0.34\textwidth}
        \centering
        \includegraphics[width=\textwidth]{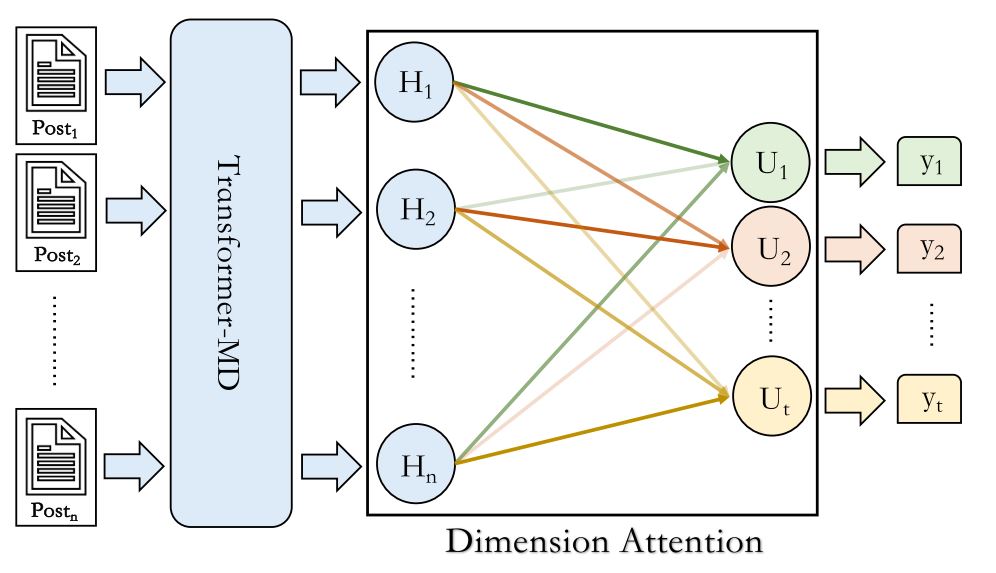}
        \caption{Overview of dimension attention module.}
    \end{subfigure}
    \caption{The schema of proposed method named Transformer-MD \cite{Yang2021}.}
    \label{fig:Yang2021}
\end{figure}

\begin{figure}
    \centering
    \includegraphics[width=\textwidth]{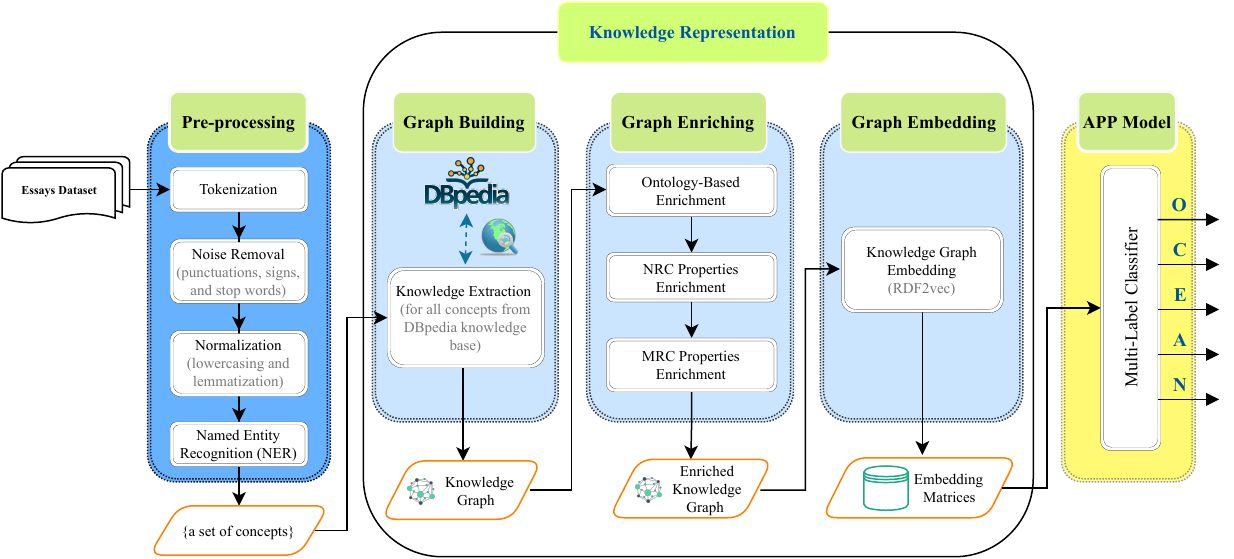}
    \caption{A knowledge graph based APP methods schemed in \cite{Ramezani_Graph-Enabled}. }
    \label{fig:ramezani-2}
\end{figure}

\begin{figure}
    \centering
    \includegraphics[width=\textwidth]{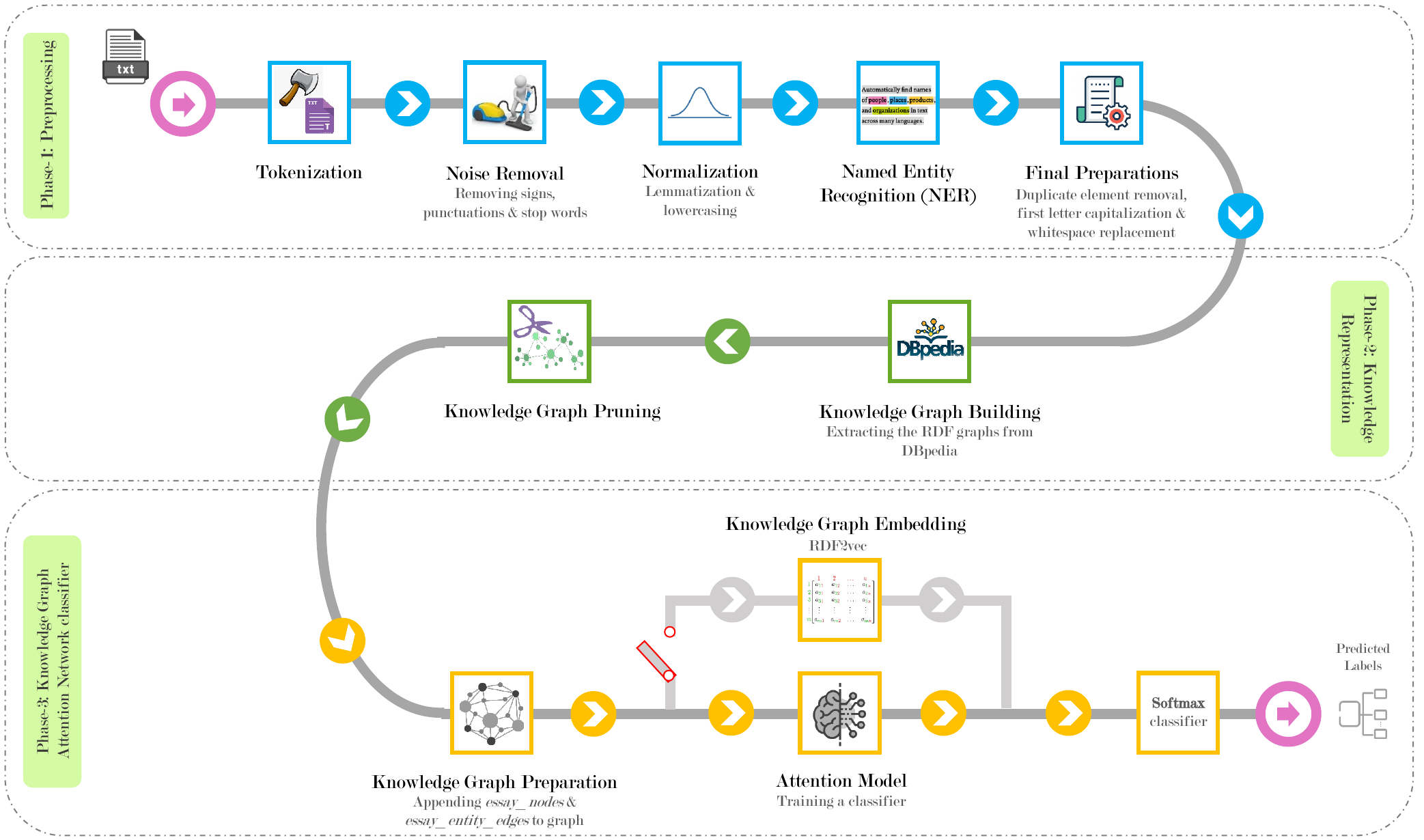}
    \caption{The general architecture of KGrAt-Net proposed by \cite{Ramezani_KGrAt-Net}. KGrAt-Net is a three-phase text classification approach which is basically founded on knowledge graph attention network.}
    \label{fig:ramezani-3}
\end{figure}

\end{document}